\documentclass[final]{cvpr}

\usepackage{times}

\usepackage{amsmath,amssymb,amsthm,amsbsy}
\usepackage{algorithm,algorithmic}
\usepackage{graphicx,subcaption}
\usepackage{booktabs,multirow,diagbox}
\usepackage{enumitem}
\usepackage{cite}
\usepackage{appendix}

\theoremstyle{remark}

\newcommand{\expect}{\mathbb{E}}

\newcommand{\nonnegativeinteger}{\mathbb{Z}^*}

\newcommand{\KL}{\mathcal{D}}
\newcommand{\CE}{\mathcal{H}}
\newcommand{\MI}{\mathcal{I}}

\usepackage[pagebackref=true,breaklinks=true,colorlinks,bookmarks=false]{hyperref}
\def\cvprPaperID{17} % *** Enter the CVPR Paper ID here
\def\confYear{CVPR 2021}

\newif\ifmain
\newif\ifsuppl

\maintrue
\suppltrue

\ifmain
\ifsuppl
\else
\fi
\else
\ifsuppl
\fi
\fi

\begin{document}

%%%%%%%%% TITLE
\title{Dual-Teacher Class-Incremental Learning With Data-Free Generative Replay}

%\author{Yoojin Choi, Mostafa El-Khamy, Jungwon Lee\\
%SoC R\&D, Samsung Semiconductor Inc., San Diego, CA 92121, USA\\
%{\tt\small \{yoojin.c,mostafa.e,jungwon2.lee\}@samsung.com}
%}
\author{%Yoojin Choi\Mark{1}, Mostafa El-Khamy\Mark{1}, Jungwon Lee\Mark{2}\\
\begin{tabular}{ccc}
Yoojin Choi, Mostafa El-Khamy & \qquad\qquad & Jungwon Lee\\
SoC R\&D, Samsung Semiconductor Inc. & & System LSI, Samsung Electronics\\
San Diego, CA 92121, USA & & South Korea\\
{\tt\small \{yoojin.c,mostafa.e\}@samsung.com} & & {\tt\small jungwon2.lee@samsung.com}\\
\end{tabular}
}

\maketitle
\thispagestyle{empty}

%%%%%%%%% ABSTRACT
\begin{abstract}
This paper proposes two novel knowledge transfer techniques for class-incremental learning (CIL). First, we propose data-free generative replay (DF-GR) to mitigate catastrophic forgetting in CIL by using synthetic samples from a generative model. In the conventional generative replay, the generative model is pre-trained for old data and shared in extra memory for later incremental learning. In our proposed DF-GR, we train a generative model from scratch without using any training data, based on the pre-trained classification model from the past, so we curtail the cost of sharing pre-trained generative models. Second, we introduce dual-teacher information distillation (DT-ID) for knowledge distillation from two teachers to one student. In CIL, we use DT-ID to learn new classes incrementally based on the pre-trained model for old classes and another model (pre-)trained on the new data for new classes. We implemented the proposed schemes on top of one of the state-of-the-art CIL methods and showed the performance improvement on CIFAR-100 and ImageNet datasets.
\end{abstract}

\section{Introduction}\label{sec:intro}

\begin{figure}[t]
\centering
\includegraphics[width=.98\columnwidth]{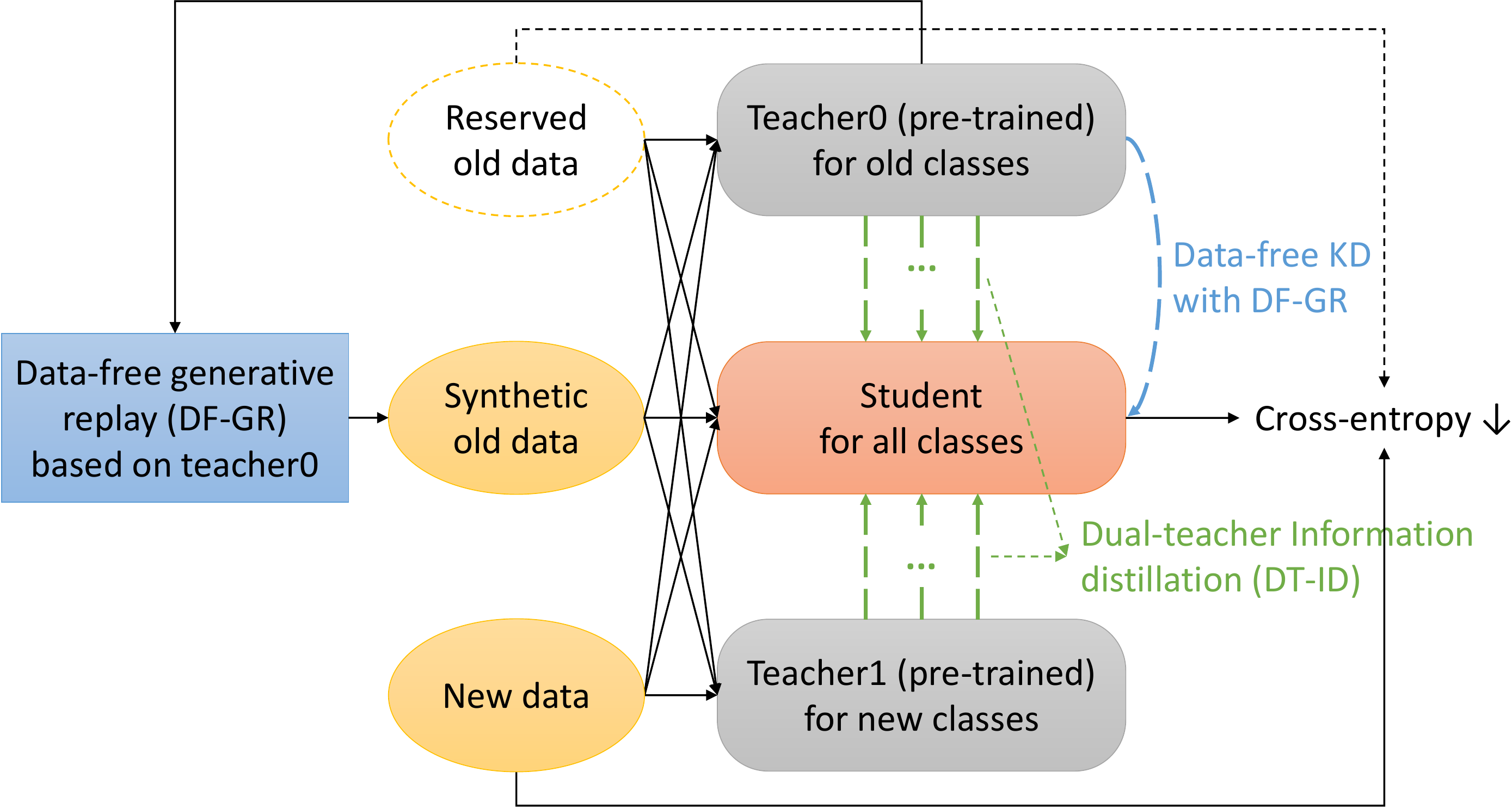}\vspace{-.5em}
\caption{Class-incremental learning with dual-teacher information distillation (DT-ID) and data-free generative replay (DF-GR). DT-ID is used to transfer knowledge from two teachers, which are trained for old and new classes, respectively, to one student. We also propose DF-GR to train a generative model for old classes from scratch without using any training data, given the pre-trained classification model for old classes. The synthetic samples from DF-GR are used for data-free KD to mitigate catastrophic forgetting in CIL.\label{sec:intro:fig:01}}\vspace{-.5em}
\end{figure}

\begin{figure*}[t]
\centering
\includegraphics[width=.9\textwidth]{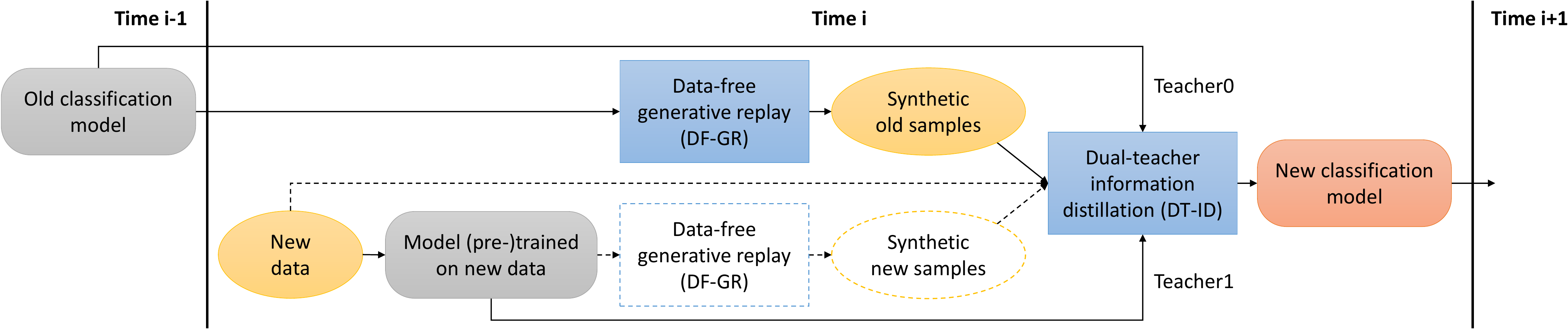}\vspace{-.5em}
\caption{Timeline of our proposed dual-teacher CIL (DT-CIL). We curtail the extra burden of storing any past data or pre-trained generative models by using DF-GR for pseudo-rehearsal of the past data. DF-GR can also be used to augment data for new classes, if not enough data are accessible for them.\label{sec:intro:fig:02}}\vspace{-.5em}
\end{figure*}

The natural learning process of humans is incremental, implying that we incrementally learn new knowledge, as we explore the world and observe new data over time. Many of real-world vision applications desire a similar incremental learning property, since their target tasks usually grow progressively by increasing demands of users, and the training data are also often collected incrementally. However, most of conventional supervised learning methods do not adapt well to this situation, since they are developed under the assumption that all the training data for learning are provided and used at once.

Incremental learning is a learning paradigm that enables a model to acquire new knowledge from new data continually, instead of training it once for all~\cite{parisi2019continual}. A baseline approach to incremental learning is to fine-tune a pre-trained model on new data when they become available, but naive fine-tuning suffers from severe performance degradation on the old tasks that the model already learned before, which is called \emph{catastrophic forgetting}~\cite{goodfellow2013empirical}. Catastrophic forgetting is caused by over-fitting to the new data when the past data are not available and cannot be used during the incremental training stages.

There have been substantial efforts devoted to mitigate catastrophic forgetting in incremental learning.

\emph{Replay methods}. Reserving some of the original training data of past tasks for future learning is a naive but effective approach to reduce catastrophic forgetting, and it was used in many of the previous works~\cite{rebuffi2017icarl,castro2018end,hou2019learning,wu2019large,liu2020mnemonics}. In \cite{wu2019large}, a subset of the reserved exemplars was used as a validation set to correct the bias towards a new task due to data imbalance. On the other hand, instead of storing raw-format samples, \emph{generative replay (GR)} proposes utilizing pre-trained generative models to reproduce synthetic samples for past tasks and use them for pseudo-rehearsal~\cite{kemker2018fearnet,shin2017continual,xiang2019incremental,liu2020generative}.

\emph{Parameter-based regularization methods}. The methods in this category mainly focus on estimating the importance of each weight in a pre-trained model and add more penalty to the distortion of the significant weights when (re-)training the model for new data~\cite{kirkpatrick2017overcoming,zenke2017continual,aljundi2018memory,nguyen2018variational,chaudhry2018riemannian}. The difference among these methods can be found in the way to compute the importance of weights. For example, in \cite{kirkpatrick2017overcoming}, the Fisher information matrix was used to measure the importance. In~\cite{nguyen2018variational}, variational continual learning introduced a prior distribution on the weights to learn the importance of the weights via a variational learning framework.

\emph{Distillation-based regularization methods}. The methods in this category utilize knowledge distillation (KD)~\cite{hinton2015distilling} to consolidate the previous knowledge in a pre-trained model, while the model is (re-)trained for new data. A distillation-based method was first introduced to incremental learning in LwF~\cite{li2017learning} to transfer the previous knowledge from a pre-trained model to a new model by matching their (softmax) outputs on new data. Then, iCaRL~\cite{rebuffi2017icarl} improved the LwF method for CIL with a small number of exemplars reserved for old classes. Moreover, iCaRL proposed nearest-mean-of-exemplars classification and prioritized construction of exemplars. The distillation-based method was improved in \cite{castro2018end} by sophisticated data augmentation with reserved samples and class-balanced fine-tuning. Recently, it was further refined in LUCIR~\cite{hou2019learning} with cosine-similarity-based classification, less-forget constraint, and inter-class separation. In \cite{dhar2019learning}, attention maps were utilized for distillation.

The main contribution of this paper is two-fold. First, we propose \emph{data-free generative replay (DF-GR)} to augment limited (old) training data with synthetic samples from a generative model. In particular, given a pre-trained model for the past tasks, the generative model is trained from scratch without using any training data, so we call it a data-free method. This is different from conventional generative replay methods~\cite{shin2017continual,xiang2019incremental,liu2020generative}, where auxiliary generative models are trained along with the main objective models by using the original training data, and the pre-trained generative models are shared for later incremental learning. Training good generative models is difficult and time-consuming. Moreover, it is redundant work in the view of current model trainers who may not care much of later incremental learning. In DF-GR, the generative model is trained by the current model trainer who wants to adapt the model for a new task, without accessing any previous data. Second, we introduce \emph{dual-teacher information distillation (DT-ID)} for knowledge distillation from two teachers to one student. In incremental learning, we use the model (pre-)trained on the new data as the second teacher for a new task, in addition to the previous one-teacher distillation method that uses the past model as the (first) teacher for old tasks. We adopt the information distillation (ID) method~\cite{ahn2019variational} and alter the original one-teacher method to a dual-teacher method. 

We investigate the usefulness of the proposed DT-ID and DF-GR methods in class-incremental learning (CIL). The goal of CIL is to learn a unified classifier that can recognize all the classes learned so far, when the training data for unseen classes arrive incrementally. Our proposed dual-teacher CIL (DT-CIL) employs both DT-ID and DF-GR, as shown in Figure~\ref{sec:intro:fig:01}. In Figure~\ref{sec:intro:fig:02}, we illustrate the timeline of DT-CIL. We implemented the proposed schemes on top of one of the state-of-the-art CIL methods, LUCIR~\cite{hou2019learning}. In our experimental results on CIFAR-100 and ImageNet datasets, we show that each proposed component provides performance improvement over the baseline LUCIR method. We also show the potential of the proposed scheme in the data-limited CIL scenario, where the available training data are limited not only for old classes, but also for new classes, due to the cost of data sharing or to preserve data privacy.

\section{Data-free generative replay (DF-GR)}\label{sec:dfgr}

Let us assume that we are given a pre-trained classification model~$t$ (called teacher) that estimates the probability distribution of class~$y$ for input~$x$. Let $g$ be the conditional generator that we train to produce the synthetic data similar to the training data used to train the teacher~$t$. The conditional generator takes a random noise vector~$z$ and a label (condition)~$y$ to produce a labeled sample. Let $p(z)$ be the random noise distribution, and let $p(y)$ be the label distribution over classes~$C$. We employ the following two losses to train the conditional generator without any training data.

\textbf{Cross-entropy (CE) loss}. We use the teacher~$t$ as a fixed discriminator to criticize the labels of the synthetic samples from the conditional generator. We define the cross-entropy loss between the label fed to the generator and the softmax output from the teacher for the generated sample as below:
\begin{equation}\label{sec:dfgr:cgen:eq:01}
L_{\text{CE}}(g|t)=\expect_{p(z)p(y)}[\CE(y,t(g(z,y)))],
\end{equation}
where $\CE$ denotes the cross-entropy, and the label~$y$ is one-hot encoded in $\CE$. 

\textbf{Batch-normalization statistics (BNS) loss}~\cite{choi2020data}. Each batch normalization (BN) layer in the pre-trained teacher~$t$ stores the mean and variance of the layer input, which we can utilize as a proxy to verify that the generator output is similar to the original training data. We use the Kullback-Leibler (KL) divergence %(e.g., see \cite[Section~2.3]{cover2012elements})
of two Gaussian distributions to match the statistics (mean and variance) stored in the BN layers of the teacher (which were obtained when trained with the original data) and the statistics computed for the generator output at the same BN layers of the teacher. Let $\mu_{l,c}$ and $\sigma_{l,c}^2$ be the mean and variance stored in batch normalization layer~$l$ of the teacher for channel~$c$. Let $\hat{\mu}_{l,c}(g)$ and $\hat{\sigma}^2_{l,c}(g)$ be the corresponding mean and variance computed based on the synthetic samples from generator~$g$. The batch-normalization statistics (BNS) loss is given by
\begin{equation}\label{sec:dfgr:cgen:eq:02}
L_{\text{BNS}}(g|t)
=\sum_{l,c}\KL_{\mathcal{N}}((\hat{\mu}_{l,c}(g),\hat{\sigma}^2_{l,c}(g)),(\mu_{l,c},\sigma_{l,c}^2)),
\end{equation}
where $\KL_{\mathcal{N}}$ is the KL divergence of two Gaussians, i.e.,
\[
\KL_{\mathcal{N}}((\hat{\mu},\hat{\sigma}^2),(\mu,\sigma^2)) \\
=\frac{(\hat{\mu}-\mu)^2+\hat{\sigma}^2}{2\sigma^2}-\log\frac{\hat{\sigma}}{\sigma}-\frac{1}{2}.
\]

\begin{figure}[t]
\centering
\includegraphics[width=\columnwidth]{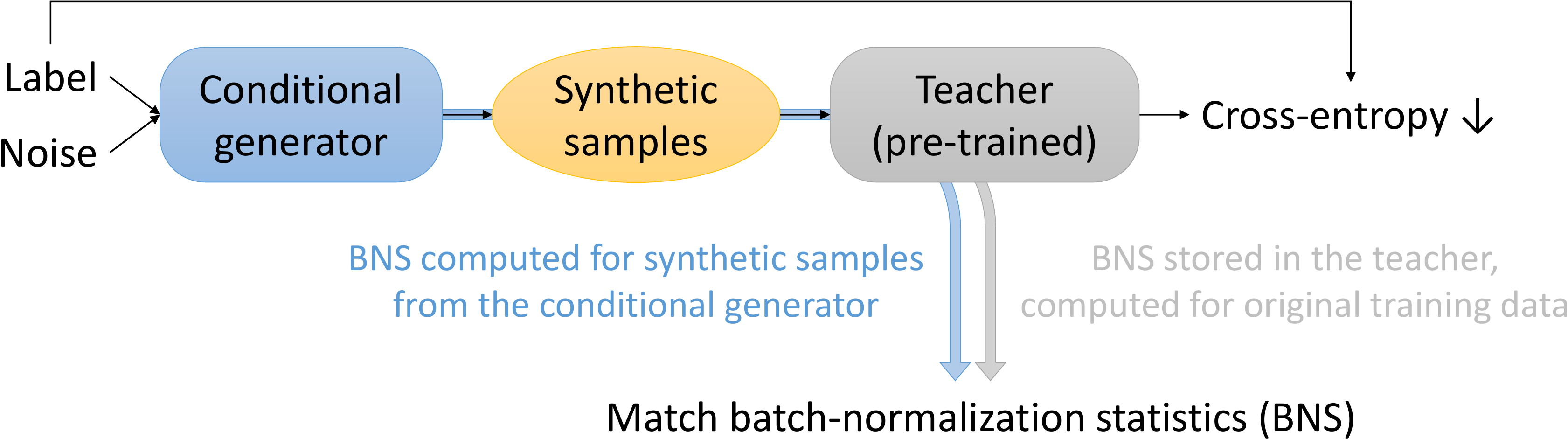}\vspace{-.5em}
\caption{Zero-shot learning of a conditional generator for DF-GR. We match the statistics stored in the batch normalization (BN) layers of the teacher (which were computed for the original data) and the statistics computed for the generator output at the same BN layers of the teacher.\label{sec:dfgr:fig:01}}\vspace{-.5em}
\end{figure}

From \eqref{sec:dfgr:cgen:eq:01} and \eqref{sec:dfgr:cgen:eq:02}, we propose the following objective for zero-shot learning of a conditional generator (see Figure~\ref{sec:dfgr:fig:01}):
\begin{equation}\label{sec:dfgr:cgen:eq:03}
\min_g\{L_{\text{CE}}(g|t)+L_{\text{BNS}}(g|t)\}.
\end{equation}
In \cite{choi2020data}, the BNS loss was introduced to train non-conditional generators for data-free KD, in particular, for data-free network quantization. In this paper, we utilize the BNS loss together with the CE loss to train a conditional generator without data. The conditional generator is then used in CIL for DF-GR, which will be elaborated in Section~\ref{sec:dtcil:dfgr}.

\setlength{\tabcolsep}{0.1em}
\begin{figure*}[t]
\centering
{\small
{\scriptsize
\begin{tabular}{cccccccccccc}
barn & cobra & cock & coral reef & cowboy hat & dam & frog & goldfish & goose & guitar & lion & meerkat\\
\includegraphics[width=.079\textwidth]{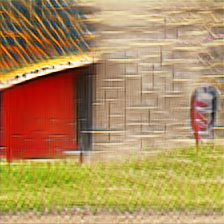}&%barn
\includegraphics[width=.079\textwidth]{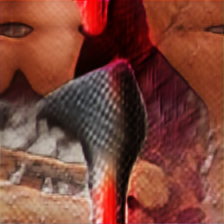}&%Indian cobra
\includegraphics[width=.079\textwidth]{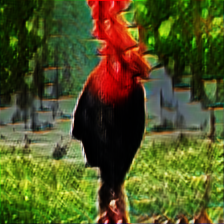}&%cock
\includegraphics[width=.079\textwidth]{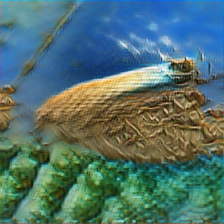}&%coral reef
\includegraphics[width=.079\textwidth]{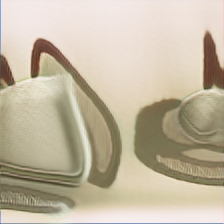}&%cowboy hat
\includegraphics[width=.079\textwidth]{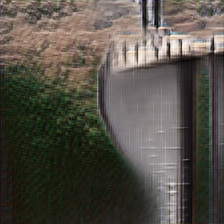}&%dam
\includegraphics[width=.079\textwidth]{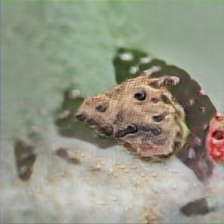}&%tailed frog, bell toad, ribbed toad, tailed toad, Ascaphus trui
\includegraphics[width=.079\textwidth]{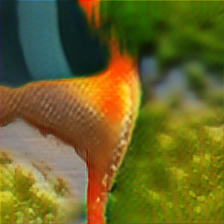}&%goldfish
\includegraphics[width=.079\textwidth]{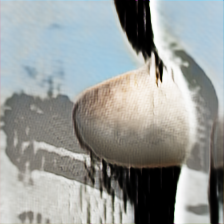}&%goose
\includegraphics[width=.079\textwidth]{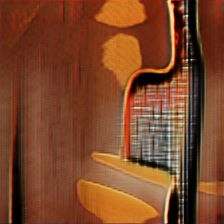}&%acoustic guitar
\includegraphics[width=.079\textwidth]{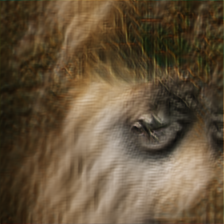}&%lion
\includegraphics[width=.079\textwidth]{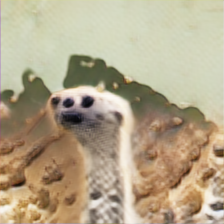}\\%meerkat
mushroom & ostrich & peacock & red wine & strawberry & submarine & trash can & violin & whiskey jug & wild boar & wolfhound & yellow orchid\\
\includegraphics[width=.079\textwidth]{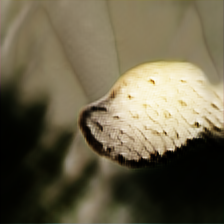}&%mushroom
\includegraphics[width=.079\textwidth]{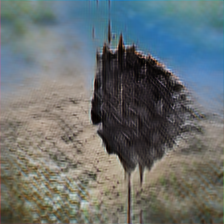}&%ostrich
\includegraphics[width=.079\textwidth]{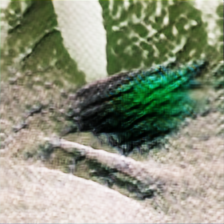}&%peacock
\includegraphics[width=.079\textwidth]{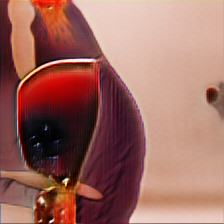}&%red wine
\includegraphics[width=.079\textwidth]{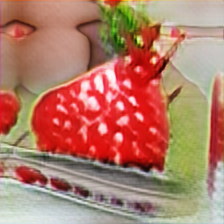}&%strawberry
\includegraphics[width=.079\textwidth]{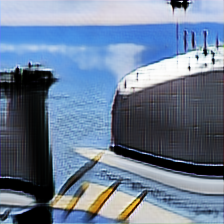}&%submarine, pigboat, sub, U-boat
\includegraphics[width=.079\textwidth]{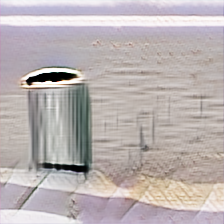}&%trash can
\includegraphics[width=.079\textwidth]{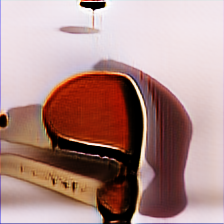}&%violin
\includegraphics[width=.079\textwidth]{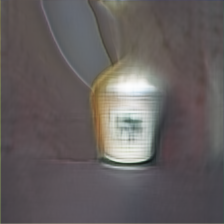}&%whiskey jug
\includegraphics[width=.079\textwidth]{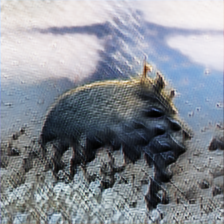}&%wild boar, boar, Sus scrofa
\includegraphics[width=.079\textwidth]{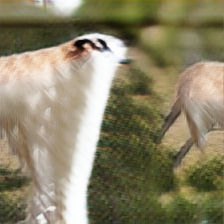}&%Russian wolfhound
\includegraphics[width=.079\textwidth]{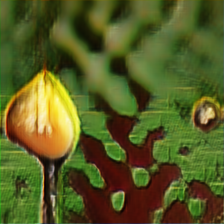}\\%yellow lady's slipper
\end{tabular}
}\\
(a) DF-GR trained with both CE and BNS losses\\\vspace{.25em}
{\scriptsize
\begin{tabular}{cccccccccccc}
barn & cobra & cock & coral reef & cowboy hat & dam & frog & goldfish & goose & guitar & lion & meerkat\\
\includegraphics[width=.079\textwidth]{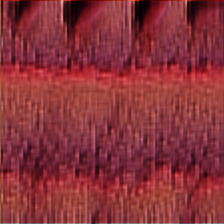}&%barn
\includegraphics[width=.079\textwidth]{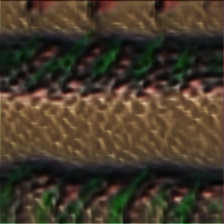}&%Indian cobra
\includegraphics[width=.079\textwidth]{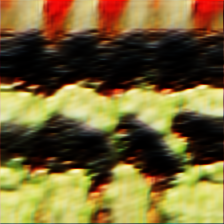}&%cock
\includegraphics[width=.079\textwidth]{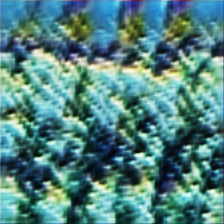}&%coral reef
\includegraphics[width=.079\textwidth]{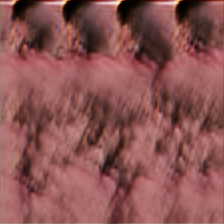}&%cowboy hat
\includegraphics[width=.079\textwidth]{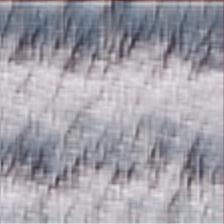}&%dam
\includegraphics[width=.079\textwidth]{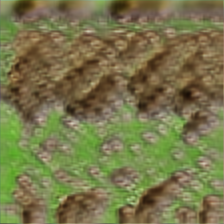}&%tailed frog, bell toad, ribbed toad, tailed toad, Ascaphus trui
\includegraphics[width=.079\textwidth]{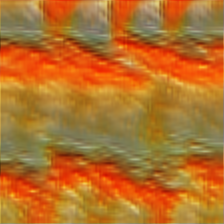}&%goldfish
\includegraphics[width=.079\textwidth]{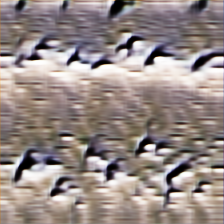}&%goose
\includegraphics[width=.079\textwidth]{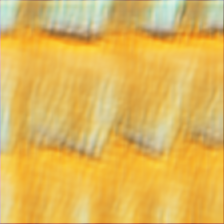}&%acoustic guitar
\includegraphics[width=.079\textwidth]{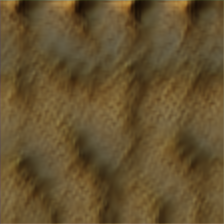}&%lion
\includegraphics[width=.079\textwidth]{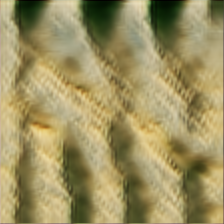}\\%meerkat
\end{tabular}
}\\
(b) DF-GR trained with CE loss only\\\vspace{.25em}
{\scriptsize
\begin{tabular}{cccccccccccc}
%barn & cobra & cock & coral reef & cowboy hat & dam & frog & goldfish & goose & meerkat\\
\includegraphics[width=.079\textwidth]{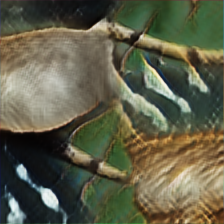}&%barn
\includegraphics[width=.079\textwidth]{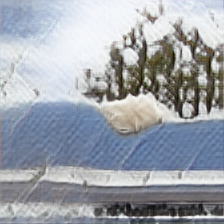}&%Indian cobra
\includegraphics[width=.079\textwidth]{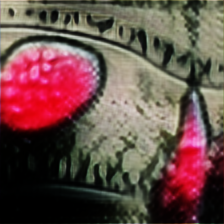}&%cock
\includegraphics[width=.079\textwidth]{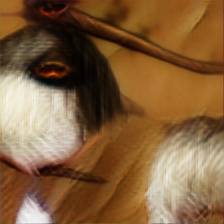}&%coral reef
\includegraphics[width=.079\textwidth]{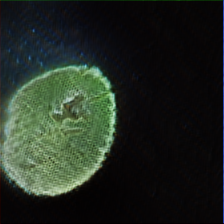}&%cowboy hat
\includegraphics[width=.079\textwidth]{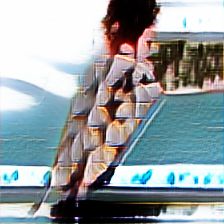}&%dam
\includegraphics[width=.079\textwidth]{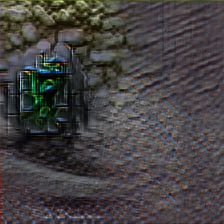}&%tailed frog, bell toad, ribbed toad, tailed toad, Ascaphus trui
\includegraphics[width=.079\textwidth]{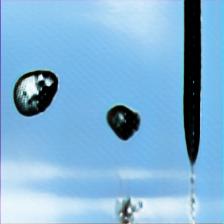}&%goldfish
\includegraphics[width=.079\textwidth]{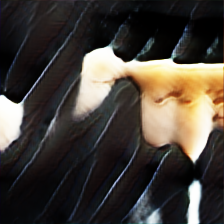}&%goose
\includegraphics[width=.079\textwidth]{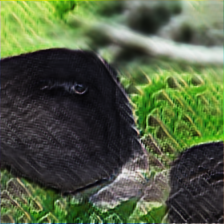}&%acoustic guitar
\includegraphics[width=.079\textwidth]{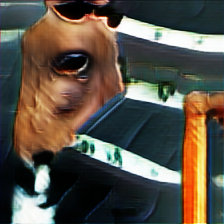}&%lion
\includegraphics[width=.079\textwidth]{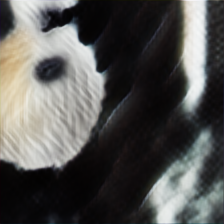}\\%meerkat
\end{tabular}
}\\
(c) DF-GR trained with BNS loss only\\
}\vspace{-.5em}
\caption{Examples of the synthetic images from DF-GR, given a pre-trained ResNet-18 for $500$ base classes from ImageNet.\label{sec:dfgr:cgen:fig:01}}
\end{figure*}

In Figure~\ref{sec:dfgr:cgen:fig:01}, we present some synthetic images from DF-GR, given a pre-trained ResNet-18 model on the ImageNet dataset as the teacher. The ResNet-18 model is pre-trained on the ImageNet data of $500$ classes that we randomly select as the base task at time~$0$ for CIL (see Section~\ref{sec:exp}). In the case of using the CE loss only, the generator fails to synthesize natural images of given classes. If we use the BNS loss only, then the conditional generator gets no feedback from the teacher (discriminator) for the label of its generated sample, so it turns into producing synthetic samples without labels. Once we utilize both CE and BNS losses, DF-GR produces labeled images that we can distinguish their classes. 

\section{Dual-teacher CIL (DT-CIL)}\label{sec:dtcil}

In CIL, we are given a sequence of classification tasks, denoted by $T_i$ for $i\in\nonnegativeinteger$, where $\nonnegativeinteger$ is the set of the non-negative integers. Task~$T_i$ at time~$i$ is a classification task for a set of classes $C_i$, such that $C_i\cap C_j=\emptyset$ for all $i\neq j$, where $\emptyset$ denotes an empty set. At time $i=0$, we train a network for (base) task~$T_0$ with base training data~$D_0$. At each time $i\geq1$, we are given a new task $T_i$ with a set of new training data~$D_i$ that belong to $C_i$, and we aim to learn the new task without forgetting the past tasks~$T_0,T_1,\dots,T_{i-1}$ that we already learned. We cannot revisit the old training data for the past tasks, unless we reserve a small number of samples, called exemplars. The reserved exemplars for task~$T_i$ are denoted by $R_i$.

Let $f_i$ be the neural network that we use at time~$i\in\nonnegativeinteger$ for CIL. Each network~$f_i$ consists of a feature extractor~$\phi_i$ and a classifier based on the extracted feature. For simplicity, we assume one-layer classifier, and let $W_0^i=\{w_c\}_{c\in C_0^i}$ be the set of classification weights used in $f_i$ for classification among all the learned classes~$C_0^i=\cup_{j=0}^iC_j$, where $W_i=\{w_c\}_{c\in C_i}$ is the classification weights newly introduced at time~$i$ for classes~$C_i$.

\subsection{Dual-teacher information distillation (DT-ID)}\label{sec:dtcil:dtid}

\setlength{\tabcolsep}{0.4em}
\begin{table}[t]
\centering
\caption{Summary of notations for dual-teacher CIL.\label{sec:cil:tbl:01}}\vspace{-.5em}
{\footnotesize
\begin{tabular}{c|cc|cc|cc|c}
\toprule
Time & \multicolumn{2}{c|}{Classes}
                           & \multicolumn{2}{c|}{Data}
                                         & \multicolumn{2}{c|}{Teachers (pre-trained)}
                                                                                         & Student\\
%\cmidrule(l{2pt}r{2pt}){2-3} \cmidrule(l{2pt}r{2pt}){4-5} \cmidrule(l{2pt}r{2pt}){6-7} \cmidrule(l{2pt}r{2pt}){8-8}
     & Old         & New   & Old*        & New   & for $C_0^{i-1}$     & for $C_i$       & for $C_0^i$\\
\midrule
$i$  & $C_0^{i-1}$ & $C_i$ & $R_0^{i-1}$ & $D_i$ & $t_0\equiv f_{i-1}$ & $t_1\equiv h_i$ & $s\equiv f_i$\\
\bottomrule
\multicolumn{8}{r}{* Reserved exemplars.}
\end{tabular}
}\vspace{-.5em}
\end{table}

The information distillation (ID) scheme~\cite{ahn2019variational} is one of the state-of-the-art KD variants, which transfers knowledge from a teacher to a student by maximizing the mutual information at their intermediate layers. We adopt this method for \emph{dual-teacher information distillation (DT-ID)} in CIL.

At time~$i$, the first teacher for DT-ID is the model~$f_{i-1}$ from time~$i-1$, which was pre-trained for old classes~$C_0^{i-1}$. We let $h_i$ be the second teacher at time~$i$, which is the model (pre-)trained on the new data at time~$i$ for new classes~$C_i$. The student is the current model~$f_i$ that we train at time~$i$ for both old and new classes in $C_0^i$. %For notational simplicity, we use the following to denote two teachers and the student, respectively, for DT-ID:
%\[
%t_0\equiv f_{i-1},
%\ \ \
%t_1\equiv h_i,
%\ \ \
%s\equiv f_i.
%\]
For notational simplicity, we use $t_0\equiv f_{i-1},t_1\equiv h_i$ and $s\equiv f_i$ to denote two teachers and the student, respectively (see Table~\ref{sec:cil:tbl:01}).

For information distillation at some of intermediate layers, we select the same $K$ intermediate layers of the teachers and the student, respectively---In practice, we select the intermediate layers with different resolutions (e.g., the layers just before down-sampling layers). Let $t_{0,k}$, $t_{1,k}$, and $s_k$ for $1\leq k\leq K$ be the feature maps from the $k$-th layer selected for DT-ID in the first teacher, the second teacher, and the student, respectively. Then, DT-ID aims to minimize
\[
L_{\text{DT-ID}}=-\sum_{k=1}^K(\MI(t_{0,k},s_k)+\MI(t_{1,k},s_k)),
\]
where $\MI$ denotes the mutual information (e.g., see \cite[Section~2.3]{cover2012elements}). As proposed in \cite{ahn2019variational}, we take the variational lower bound of the mutual information and perform variational information maximization~\cite{barber2003algorithm} with a Gaussian prior. Then, for each $n\in\{0,1\}$, it can be shown that
\begin{equation}\label{sec:dtcil:dtid:eq:01}
-\MI(t_{n,k},s_k)
\leq
%-\tilde{\MI}(t_{n,k},s_k)
%\triangleq
\expect[\mathcal{J}_{n,k}(t_{n,k},s_k)]+X,
\end{equation}
where
\begin{equation}\label{sec:dtcil:dtid:eq:02}
\mathcal{J}_{n,k}(t,s)
=\sum_{c,h,w}\frac{(t_{c,h,w}-\mu_{k,c,h,w}^n(s))^2}{2\sigma_{k,c}^2}+\log\sigma_{k,c},
\end{equation}
for some constant~$X$; $t_{c,h,w}$ is the scalar element of tensor~$t$ at channel~$c$, height~$h$, width~$w$, and $\mu_{k,c,h,w}^n(s)$ is the output of the neural network~$\mu_k^n$ at channel~$c$, height~$h$, width~$w$, when it takes $s$ as its input---$\mu_k^n$ is the convolutional network used to transform the student feature maps into the teacher domain at each intermediate layer~$k$ selected for ID. We propose using a common variance~$\sigma_{k,c}^2$ for both $n\in\{0,1\}$ at each layer~$k$ and channel~$c$, so we transfer information from two teachers without inclining towards either of them. We note that the detailed steps to derive \eqref{sec:dtcil:dtid:eq:01} and \eqref{sec:dtcil:dtid:eq:02} from variational information maximization can be found in \cite{ahn2019variational,barber2003algorithm} and are omitted here due to the page limit. 

\textbf{Dual-teacher information distillation (DT-ID) loss}. In practice, we take empirical expectation in \eqref{sec:dtcil:dtid:eq:01} by using the feature maps obtained from all available data at time~$i$ and define the DT-ID loss as below, for $A_i=D_i\cup R_0^{i-1}$,
\begin{multline}\label{sec:dtcil:dtid:eq:03}
\bar{L}_{\text{DT-ID}}^i=\sum_{k=1}^K\expect_{(x,y)\in A_i}
[\mathcal{J}_{0,k}(f_{i-1,k}(x),f_{i,k}(x))\\
+\mathcal{J}_{1,k}(h_{i,k}(x),f_{i,k}(x))].
\end{multline}

For the baseline CIL scheme, we adopt three key components of LUCIR~\cite{hou2019learning}, i.e., cosine-similarity-based classifier, cross-entropy loss, and less-forget constraint.

\textbf{Cosine-similarity-based classifier}. In \cite{hou2019learning}, it was proposed for the classifier to use the cosine similarity instead of the conventional dot-product, to resolve the data imbalance problem, when only a small number of reserved exemplars are given for old classes. The cosine-similarity-based classifier yields, before applying the softmax function,
\begin{equation}\label{sec:cil:eq:01}
l_{i,c}(x)=\left<w_c,\phi_i(x)\right>=\frac{w_c^T\phi_i(x)}{|w_c||\phi_i(x)|},
\ \ \
c\in C_0^i,
\end{equation}
for input $x$, where $a^T$ is the transpose of $a$, $|a|$ is the $l^2$-norm of $a$, and $\left<a,b\right>$ is the cosine similarity between $a$ and $b$. The network~$f_i$ computes the categorical probability distribution of input~$x$ over classes~$C_0^i$ by encoding the cosine-similarity scores with the softmax function.%, i.e., 
%\[
%f_i(x)=\softmax(\{l_{i,c}(x)\}_{c\in C_0^i}).
%\]

\textbf{Cross-entropy (CE) loss}. The categorical cross-entropy loss is computed on the labeled training data for new classes and the reserved exemplars, if present, for old classes:
\begin{equation}\label{sec:cil:eq:02}
L_{\text{CE}}^i=\expect_{(x,y)\in A_i}[\CE(y,f_i(x))],\ \ \ A_i=D_i\cup R_0^{i-1},
\end{equation}
where $(x,y)$ is the pair of a training sample and its ground-truth label.%, $\CE$ denotes the cross-entropy, and the label~$y$ is one-hot encoded in $\CE$.

\textbf{Less-forget (LF) constraint}~\cite{hou2019learning}. To maintain the previous knowledge learned in a pre-trained model, it was proposed to inherit and fix the weights for old classes from the pre-trained model and minimize the following LF loss:
\begin{equation}\label{sec:cil:eq:03}
L_{\text{LF}}^i=-\expect_{(x,y)\in A_i}[\left<\phi_{i-1}(x),\phi_i(x)\right>].
\end{equation}
The LF loss constrains the change of the feature extractor, when (re-)training the model on new data so that it does not forget the knowledge for old classes.

By combining the DT-ID, CE and LF losses in \eqref{sec:dtcil:dtid:eq:03}, \eqref{sec:cil:eq:02}, and \eqref{sec:cil:eq:03}, the final objective of our DT-CIL is given by
\begin{equation}\label{sec:dtcil:dtid:eq:04}
\min_{\phi_i,W_i,\mu\text{'s},\sigma^2\text{'s}}\{\bar{L}_{\text{DT-ID}}^i+L_{\text{CE}}^i+\alpha_i L_{\text{LF}}^i\},
\end{equation}
where $\alpha_i\geq0$ is the factor to control less-forgetting.

\subsection{DT-CIL with DF-GR}\label{sec:dtcil:dfgr}

For DF-GR at time~$i$ of CIL, the pre-trained model~$f_{i-1}$ for old classes from time~$i-1$ is set to be the teacher~$t$ in \eqref{sec:dfgr:cgen:eq:03}, and a conditional generator~$g=g_i$ is trained to reproduce the synthetic samples for old classes~$C_0^{i-1}$ by \eqref{sec:dfgr:cgen:eq:03}. We emphasize that at each time~$i$, we train a new generator~$g_i$, from scratch without using any training data, based on the previous model~$f_{i-1}$, and no pre-trained generators are transferred to the future (see Figure~\ref{sec:intro:fig:02}). The synthetic samples from $g_i$ are used in CIL as follows:
\begin{itemize}[noitemsep,topsep=0em,leftmargin=1.2em]
\item First, we include the synthetic samples when computing the DT-ID and LF losses in \eqref{sec:dtcil:dtid:eq:03} and \eqref{sec:cil:eq:03}, i.e., we add the synthetic samples to the available data~$A_i$ in \eqref{sec:dtcil:dtid:eq:03} and \eqref{sec:cil:eq:03}.
\item Second, we add the data-free KD loss (see \eqref{sec:dtcil:dfgr:eq:02}) for the synthetic samples to our DT-CIL objective in \eqref{sec:dtcil:dtid:eq:04}, which is elaborated below.
\item We do not compute the CE loss in \eqref{sec:cil:eq:02} with the synthetic samples, since the ground-truth labels for the synthetic samples are unknown although they are machine-labeled in conditional generation.
\end{itemize}

\textbf{Data-free KD with weight imprinting}. Note that the past model~$f_{i-1}$ yields the softmax output for old classes only, while the current model~$f_i$ that we train at time~$i$ produces the softmax output for both old and new classes. For KD from $f_{i-1}$ to $f_i$, we need to match the number of output classes. In \cite{li2017learning,wu2019large}, it was proposed to match the softmax output for old classes only in KD. In this paper, we propose extending the classification layer of $f_{i-1}$ to cover the new classes by \emph{weight imprinting}~\cite{qi2018low} and matching the softmax output for both old and new classes. We will show the gain of weight imprinting in our experiments (see Figure~\ref{sec:exp:fig:05}).

For weight imprinting, we collect the output of the feature extractor~$\phi_{i-1}$ for every training sample of each new class and use their average as the classification weight of that class. If we use a cosine-similarity-based classifier (see \eqref{sec:cil:eq:01}), we normalize the features before taking their average:
\begin{equation}\label{sec:dtcil:dfgr:eq:01}
w_c=\expect_{(x,y)\in D_i(c)}\left[\frac{\phi_{i-1}(x)}{|\phi_{i-1}(x)|}\right],
\ \ \
c\in C_i,
\end{equation}
where $D_i(c)$ is the samples of class $c\in C_i$ in $D_i$. In \eqref{sec:dtcil:dfgr:eq:01}, weight imprinting basically finds the weight that maximizes the average cosine similarity to the features extracted from the available training samples for each class. 

Let $\tilde{f}_{i-1}$ be the past model with an extended classifier after weight imprinting for new classes. Then, the data-free KD (DF-KD) loss using DF-GR is given by
\begin{multline}\label{sec:dtcil:dfgr:eq:02}
L_{\text{DF-KD}}^i \\
=\expect_{p(z)p_i(y)}[\CE(\tilde{f}_{i-1}(g_i(z,y)),f_i(g_i(z,y)))],
\end{multline}
where $p_i(y)$ is the label distribution, for which we use the uniform distribution over all the past classes~$C_0^{i-1}$ at time~$i$. The DF-KD loss is finally added to \eqref{sec:dtcil:dtid:eq:04} for DT-CIL.

\section{Experiments}\label{sec:exp}

We perform experiments on our proposed DT-CIL with DT-ID and DF-GR for CIFAR-100~\cite{krizhevsky2009learning} and ImageNet~\cite{russakovsky2015imagenet} 2012 datasets. For ImageNet, we consider two cases. First, we evaluate CIL on $100$ classes randomly selected from the full ImageNet data, which is called ImageNet-Subset. Second, we evaluate CIL on all $1000$ classes of the ImageNet dataset, which is called ImageNet-Full. For CIL on CIFAR-100 and ImageNet-Subset, we split the classes into $6$ tasks, where the base task at time~$0$ has $50$ classes, and each of the following $5$ tasks for time~$i\in\{1,2,3,4,5\}$ has $10$ classes. For ImageNet-Full, we split the classes into the base task of $500$ classes and the following $5$ tasks of $100$ classes in each.

\subsection{Implementation details}\label{sec:exp:impl}

We use 32-layer and 18-layer ResNets~\cite{he2016deep} (ResNet-32 and ResNet-18) for CIFAR-100 and ImageNet, respectively. We adopt the cosine-similarity-based classifier in \eqref{sec:cil:eq:01}, while we remove the last rectified linear unit (ReLU) from the original ResNet feature extractor, so the features can be any real values. We mostly follow the training hyper-parameters suggested in LUCIR~\cite{hou2019learning}, on top of which we additional implement our DT-ID and DF-GR methods. Due to the page limit, we defer more details to the supplementary materials.

For the conditional generator of DF-GR, we adopt the U-Net~\cite{ronneberger2015u} architecture used in \cite{ulyanov2018deep}. We replace batch normalization with conditional batch normalization~\cite{de2017modulating,miyato2018cgans} for conditional generation. To train the conditional generator for \eqref{sec:dfgr:cgen:eq:03}, we employ Adam optimizer~\cite{kingma2014adam} with momentum~$0.5$ and learning rate~$10^{-3}$. We use $80$ and $45$ epochs of batch size $256$ and $100$ for CIFAR-100 and ImageNet, respectively, where each epoch has $50$, $128$, and $1280$ batches for CIFAR-100, ImageNet-Subset, and ImageNet-Full, respectively. The trained generator is then used to compute the DF-KD loss in \eqref{sec:dtcil:dfgr:eq:02}. In each batch to compute the DF-KD loss, we generate new synthetic samples, while we continue updating the generator for \eqref{sec:dfgr:cgen:eq:03}.

For DT-ID, we choose the layers at the end of each group of residual blocks, except the first group. Each network~$\mu_k^n$ in \eqref{sec:dtcil:dtid:eq:02} consists three convolutional layers of kernel size~$1\times1$, where we double the channel size in the first two convolutional layers and reduce it back to the original size at the last convolution layer to compare its output to the feature map from the teacher. Batch normalization and ReLU activation are included in the first two convolutional layers. To ensure $\sigma_{k,c}^2$ to be always positive, we use a proxy~$\omega_{k,c}=\log{\sigma_{k,c}^2}$ for gradient descent. The initial value of $\omega_{k,c}$ is set to be $0$.

\subsection{Evaluation metrics}\label{app:exp:metric}

\emph{Average accuracy}. Let $A_i(c)$ for $c\in C_0^i$ be the accuracy on the test data of class~$c$ at time~$i$. The average accuracy at time~$i$ is the average of $A_i(c)$ for all seen classes~$c\in C_0^i$.
%\[
%A_i=\frac{1}{|C_0^i|}\sum_{c\in C_0^i}A_i(c).
%\]

\emph{Average forgetting}~\cite{chaudhry2018riemannian}. %The average forgetting was suggested in \cite{chaudhry2018riemannian} to measure the level of forgetting the previous knowledge for old classes.
The forgotten knowledge for an old class~$c\in C_0^{i-1}$ at time $i\geq1$ is measured by
\[
F_i(c)=\max_{j\leq k\leq i-1}\{A_k(c)-A_i(c)\},
\ \ \
c\in C_j,
\ \ \
0\leq j\leq i-1,
\]
and the average forgetting at time~$i$ is the average of $F_i(c)$ for all old classes~$c\in C_0^{i-1}$. %as below:
%\[
%F_i=\frac{1}{|C_0^{i-1}|}\sum_{c\in C_0^{i-1}}F_i(c).
%\]
%The average forgetting gives an estimate of how much the model forgot about the class given its current state. 

\subsection{Experimental results}\label{sec:exp:res}

\setlength{\tabcolsep}{0.25em}
\begin{table}[t]
\centering
\caption{Notations used in experimental results.\label{sec:exp:tbl:01}}\vspace{-.5em}
{\footnotesize
\begin{tabular}{cll}
\toprule
$N_D$ & \multicolumn{2}{l}{\# training data (or reserved exemplars) for each new class}\\
\midrule
$N_R$ & \multicolumn{2}{l}{\# reserved exemplars for each old class}\\
\midrule
$N_t$ & \# teachers,
%\cmidrule{2-2}
       & $1$: The baseline one-teacher method\\
&      & $2$: The proposed dual-teacher method using DT-ID\\
\midrule
$N_g$ & \# generators,
%\cmidrule{2-2}
       & $0$: No DF-GR\\
&      & $1$: Single DF-GR for old classes\\
&      & $2$: Dual DF-GR for both old and new classes\\
\bottomrule
\end{tabular}
}
\end{table}

\setlength{\tabcolsep}{0.2em}
\begin{figure}[t]
\centering
{\small
\includegraphics[scale=.45]{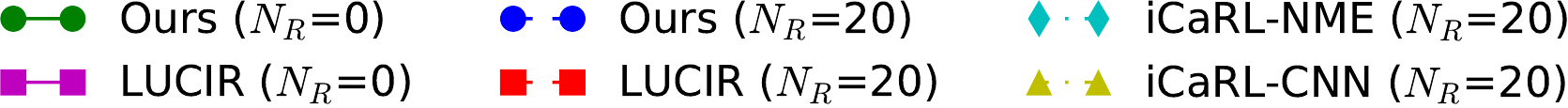}\\\vspace{.5em}
\begin{tabular}{cc}
\includegraphics[height=.38\columnwidth]{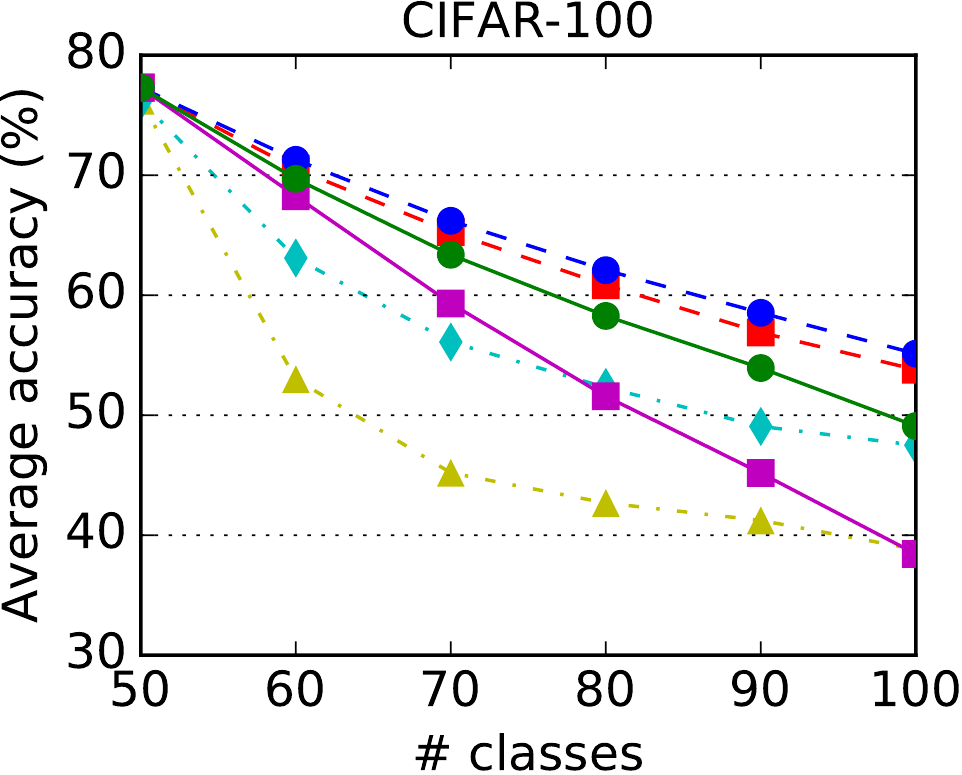}&
\includegraphics[height=.38\columnwidth]{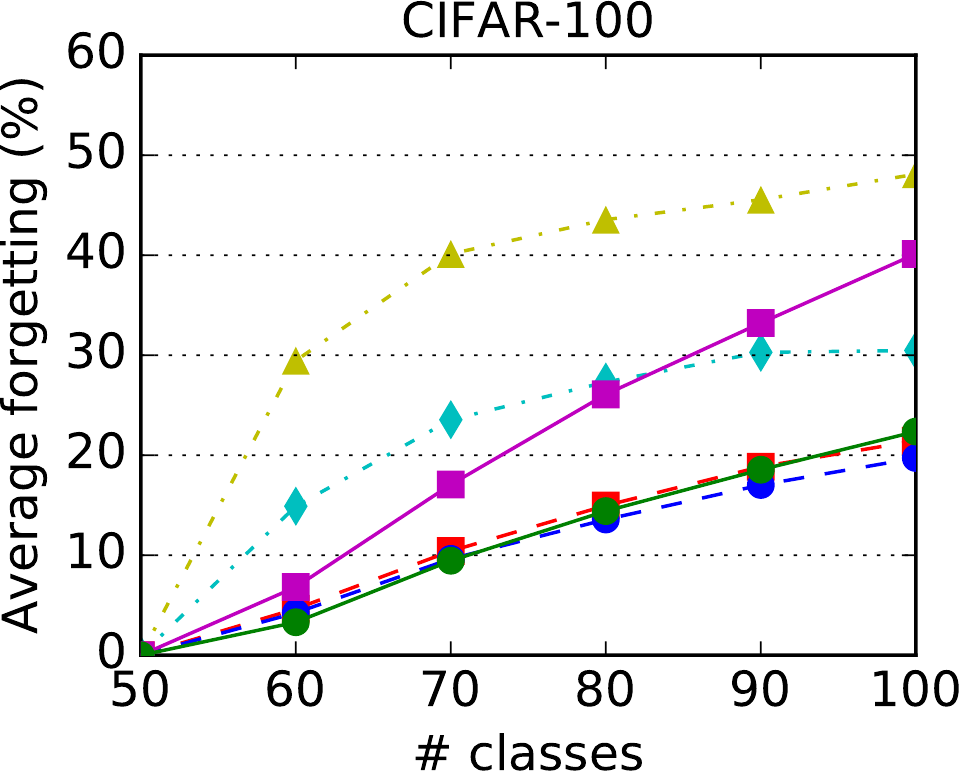}
\end{tabular}
%(a) CIFAR-100\\\vspace{.5em}
%\includegraphics[height=.06\columnwidth]{figs/model0resnet18toresnet18cosimagenet2012sublgd-crop}\\\vspace{.25em}
\begin{tabular}{cc}
\includegraphics[height=.38\columnwidth]{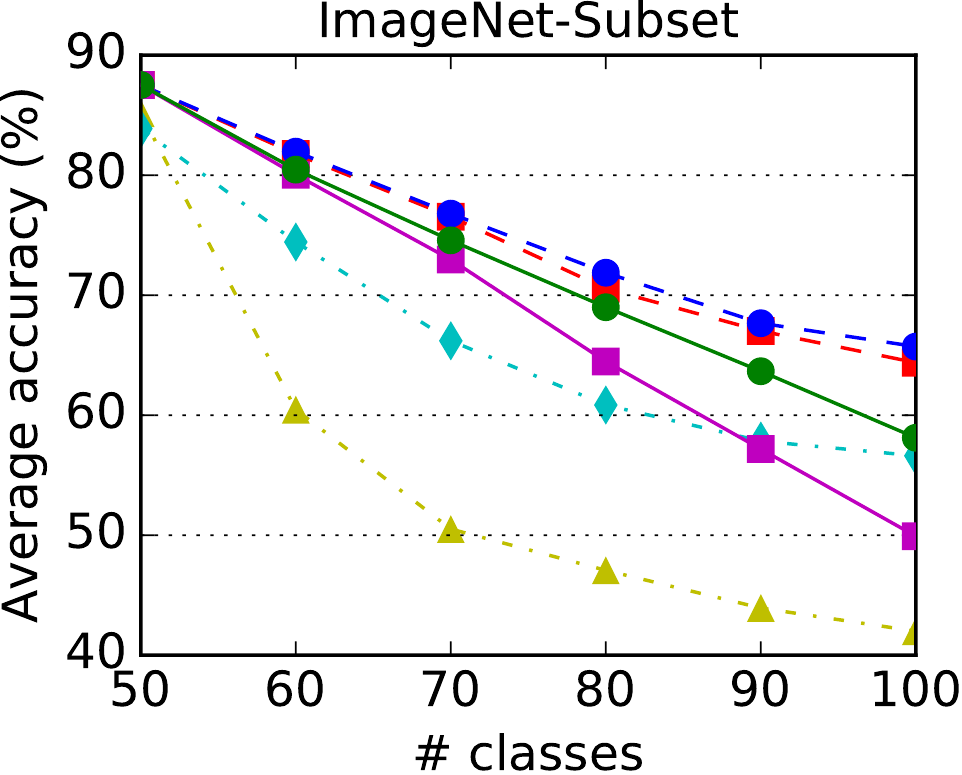}&
\includegraphics[height=.38\columnwidth]{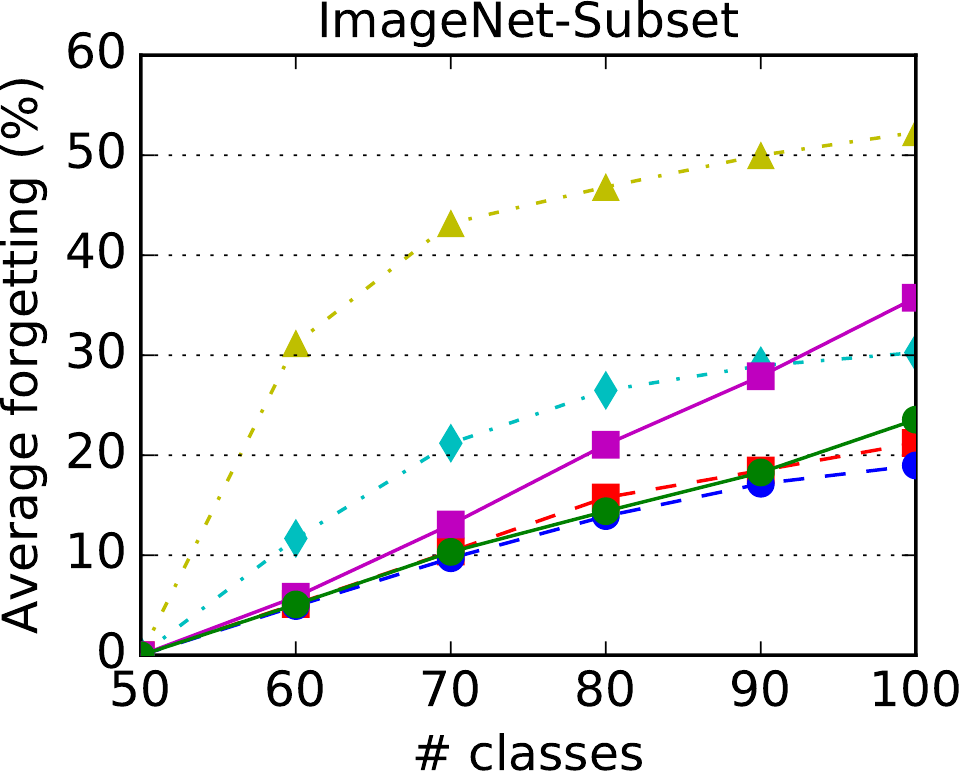}
\end{tabular}
%(b) ImageNet-Subset\\\vspace{.5em}
%\includegraphics[height=.06\columnwidth]{figs/model0resnet18toresnet18cosimagenet2012lgd-crop}\\\vspace{.25em}
\begin{tabular}{cc}
\includegraphics[height=.38\columnwidth]{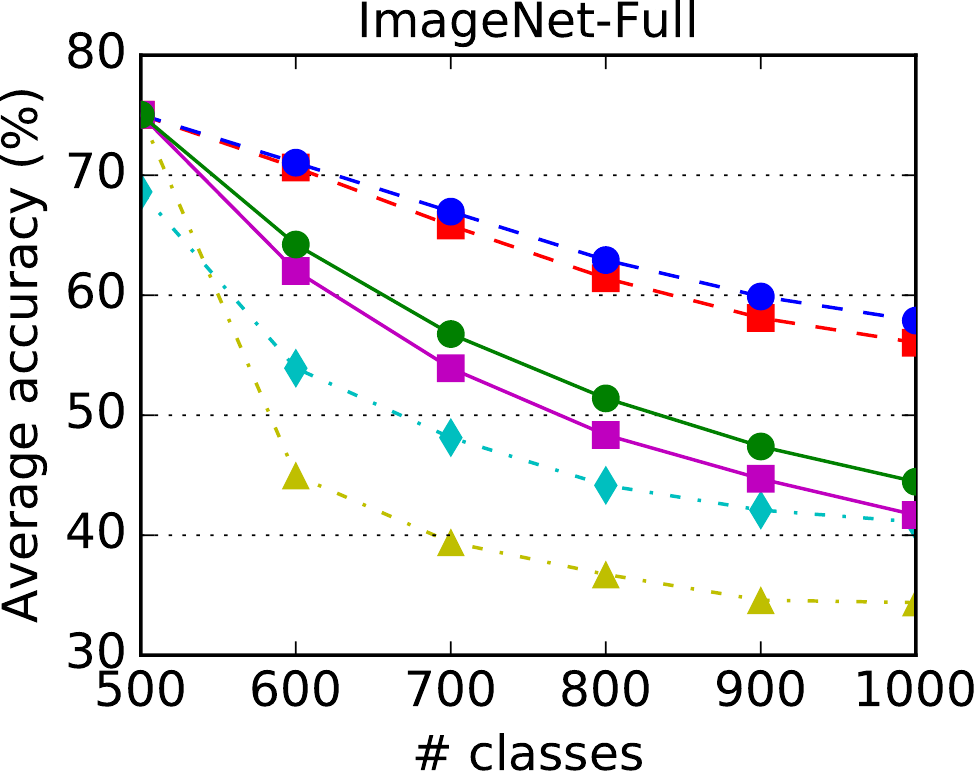}&
\includegraphics[height=.38\columnwidth]{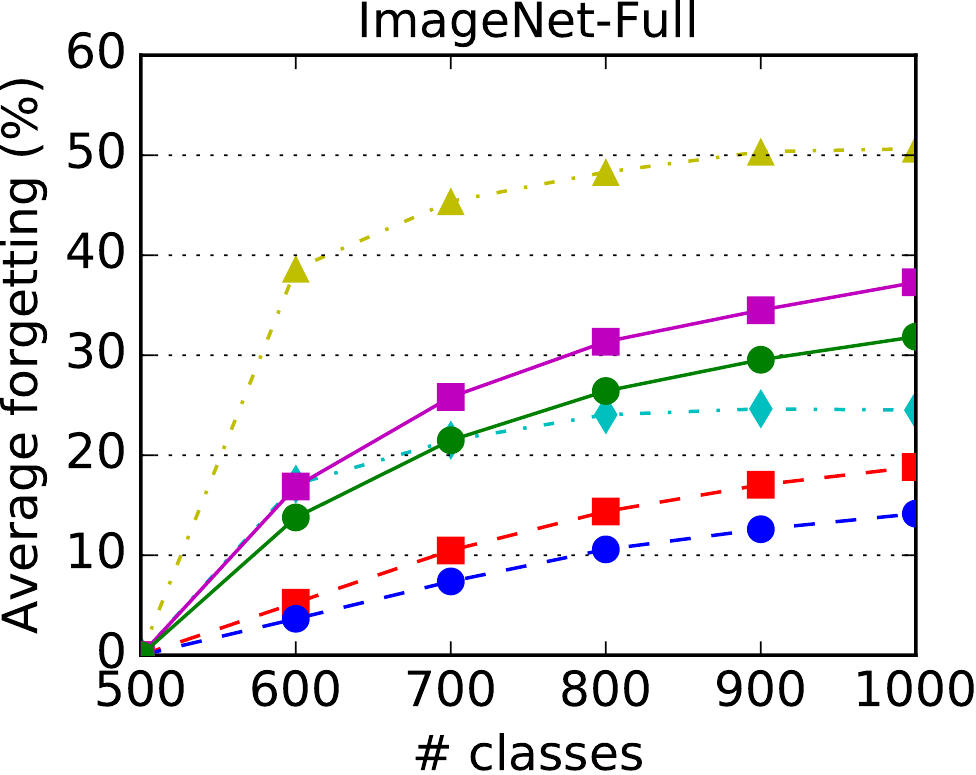}
\end{tabular}
%(c) ImageNet-Full\\
}\vspace{-.5em}
\caption{Average accuracy and average forgetting in CIL on CIFAR-100, ImageNet-Subset, and ImageNet-Full. For ours, we used both DT-ID and DF-GR ($N_t,N_g=2,1$).\label{sec:exp:fig:01}}
\end{figure}

First, Table~\ref{sec:exp:tbl:01} summarizes the notations used in our experimental results. In Figure~\ref{sec:exp:fig:01}, we show the results of CIL on CIFAR-100, ImageNet-Sub, and ImageNet-Full, in the conventional CIL scenario, where the training data for new classes in the datasets are all available. In particular, we evaluate two cases of $N_R=0$ and $N_R=20$. For the results of ours in Figure~\ref{sec:exp:fig:01}, we use both DT-ID and DF-GR, i.e., $N_t,N_g=2,1$. We compare ours with the baseline LUCIR~\cite{hou2019learning} and iCaRL~\cite{rebuffi2017icarl}. For iCaRL, we followed the implementation in \cite{hou2019learning}, which is slightly different but better than the original one. We show the results for iCaRL in two cases when we use the nearest-mean-of-exemplars classifier (iCaRL-NME) and when we use the softmax scores for classification (iCaRL-CNN). In Figure~\ref{sec:exp:fig:01}, observe that our proposed DT-CIL method with DT-ID and DF-GR provides gains in accuracy over the baseline LUCIR~\cite{hou2019learning}. In particular, the gain is significant in case that we have no reserved old exemplars ($N_R=0$). From the plots for average forgetting, one can see that the proposed methods help to reduce catastrophic forgetting when reserved old exemplars are limited.

\setlength{\tabcolsep}{0.45em}
\begin{table*}[t]
\centering
\caption{Average accuracy and average forgetting at each time of CIL on ImageNet-Full in the conventional CIL scenario.\label{sec:exp:tbl:02}}\vspace{-.5em}
{\footnotesize
\begin{tabular}{ccccccccc|cccccc}
\toprule
& $N_D,N_R$ & $N_t,N_g$ & \multicolumn{6}{c}{Average accuracy (\%) for all seen classes} & \multicolumn{6}{|c}{Average forgetting (\%) for old classes}\\
\midrule
%\cmidrule(l{2pt}r{2pt}){4-9} \cmidrule(l{2pt}r{2pt}){10-15}
Time
&           &           & 0 & 1 & 2 & 3 & 4 & 5 & 0 & 1 & 2 & 3 & 4 & 5\\
(\# classes)       
&           &           & (500) & (+100) & (+100) & (+100) & (+100) & (+100) & (500) & (+100)  & (+100) & (+100) & (+100) & (+100)\\
\midrule
%LwF.MC
%& 1.3k, 0   & 1, 0      & 75.03 & 14.83 & 12.76 & 11.12 & 9.93 & 9.29 & 0.00 & 74.99 & 76.69 & 77.96 & 78.69 & 79.21\\
LUCIR~\cite{hou2019learning}*
& 1.3k, 0   & 1, 0      & 75.04 & 62.02 & 53.91 & 48.36 & 44.70 & 41.68 & 0.00 & 16.86 & 25.82 & 31.35 & 34.51 & 37.29\\
\midrule
\multirow{3}{*}{Ours}
&1.3k, 0    & 2, 0      & 75.04 & 62.62 & 54.63 & 49.05 & 45.11 & 41.83 & 0.00 & 16.02 & 24.83 & 30.29 & 33.69 & 36.73\\
&1.3k, 0    & 1, 1      & 75.04 & 63.41 & 55.71 & 50.13 & 46.12 & 43.01 & 0.00 & 14.70 & 22.70 & 27.83 & 31.03 & 33.66\\
&1.3k, 0    & 2, 1      & 75.04 & \textbf{64.23} & \textbf{56.77} & \textbf{51.40} & \textbf{47.40} & \textbf{44.46} & 0.00 & \textbf{13.78} & \textbf{21.50} & \textbf{26.42} & \textbf{29.55} & \textbf{31.86}\\
\midrule
%iCaRL-CNN
%&1.3k, 20   & 1, 0      & 75.03 & 44.98 & 39.41 & 36.73 & 34.59 & 34.39 & 0.00 & 38.64 & 45.39 & 48.32 & 50.35 & 50.65\\
%iCaRL-NME
%&1.3k, 20   & 1, 0      & 68.61 & 53.92 & 48.13 & 44.16 & 42.09 & 41.11 & 0.00 & 17.10 & 21.50 & 24.05 & 24.64 & 24.51\\
LUCIR~\cite{hou2019learning}*
&1.3k, 20   & 1, 0      & 75.04 & 70.64 & 65.81 & 61.42 & 58.10 & 56.05 & 0.00 & 5.25 & 10.48 & 14.38 & 17.04 & 18.83\\
\midrule
\multirow{3}{*}{Ours}
&1.3k, 20   & 2, 0      & 75.04 & 70.71 & 66.41 & 62.20 & 59.18 & 57.04 & 0.00 & 4.72 & 8.85 & 12.43 & 14.74 & 16.58\\
&1.3k, 20   & 1, 1      & 75.04 & 70.92 & 66.24 & 62.39 & 59.33 & 57.07 & 0.00 & 4.06 & 8.53 & 11.67 & 14.02 & 15.76\\
&1.3k, 20   & 2, 1      & 75.04 & \textbf{71.05} & \textbf{66.97} & \textbf{62.93} & \textbf{59.90} & \textbf{57.89} & 0.00 & \textbf{3.64} & \textbf{7.38} & \textbf{10.58} & \textbf{12.60} & \textbf{14.16}\\
\bottomrule
\multicolumn{15}{r}{* Reproduced by us.}
\end{tabular}
}
\end{table*}

Table~\ref{sec:exp:tbl:02} provides the results of an ablation study for each component, DT-ID and DF-GR, of our proposed method for CIL on ImageNet-Full. First, $N_t,N_g=1,0$ implies the baseline LUCIR~\cite{hou2019learning}. Second, using DT-ID on top of LUCIR corresponds to the case of $N_t,N_g=2,0$. Third, using DF-GR with LUCIR is denoted by $N_t,N_g=1,1$. Finally, in the case of $N_t,N_g=2,1$, we use both DT-ID and DF-GR (see Table~\ref{sec:exp:tbl:01} for the notations). Table~\ref{sec:exp:tbl:02} shows that each proposed component provides performance improvement over the baseline in both cases when we have a small number of reserved exemplars ($20$ per each old class) and when we have no reserved exemplars for old classes.

\setlength{\tabcolsep}{0.2em}
\begin{figure}[t]
\centering
{\small
\includegraphics[scale=.45]{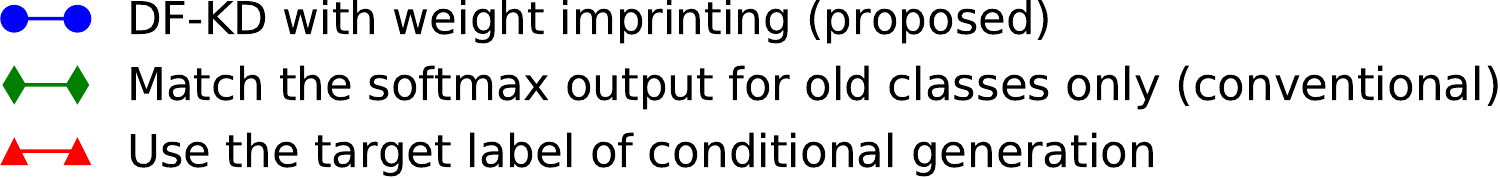}\\\vspace{.5em}
\begin{tabular}{cc}
\includegraphics[height=.37\columnwidth]{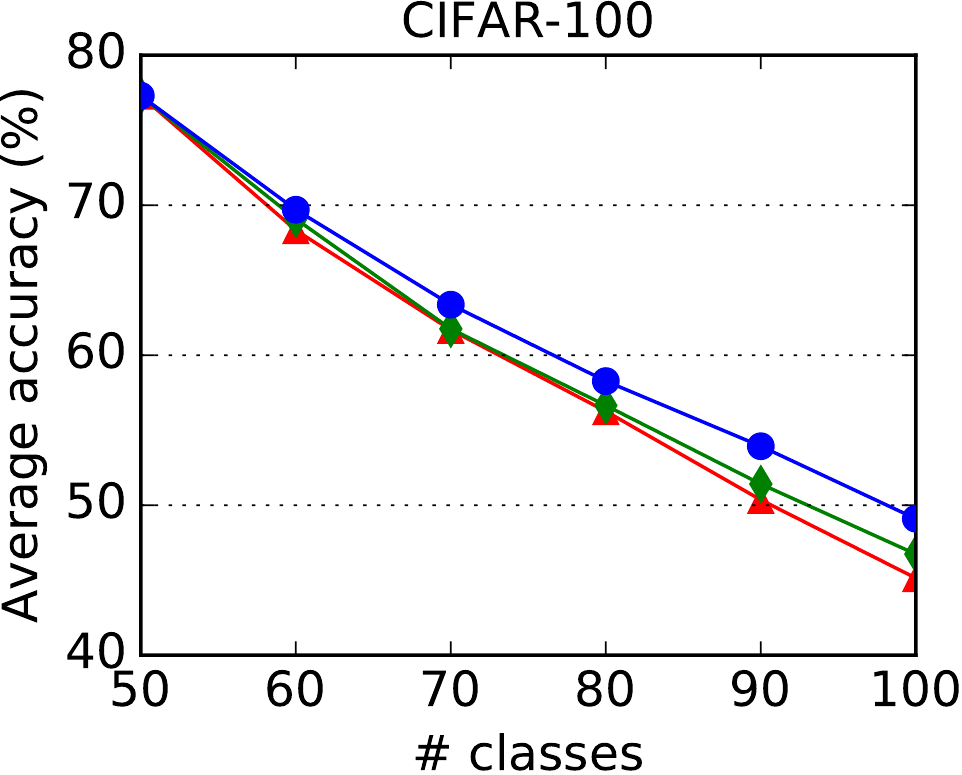}&
\includegraphics[height=.37\columnwidth]{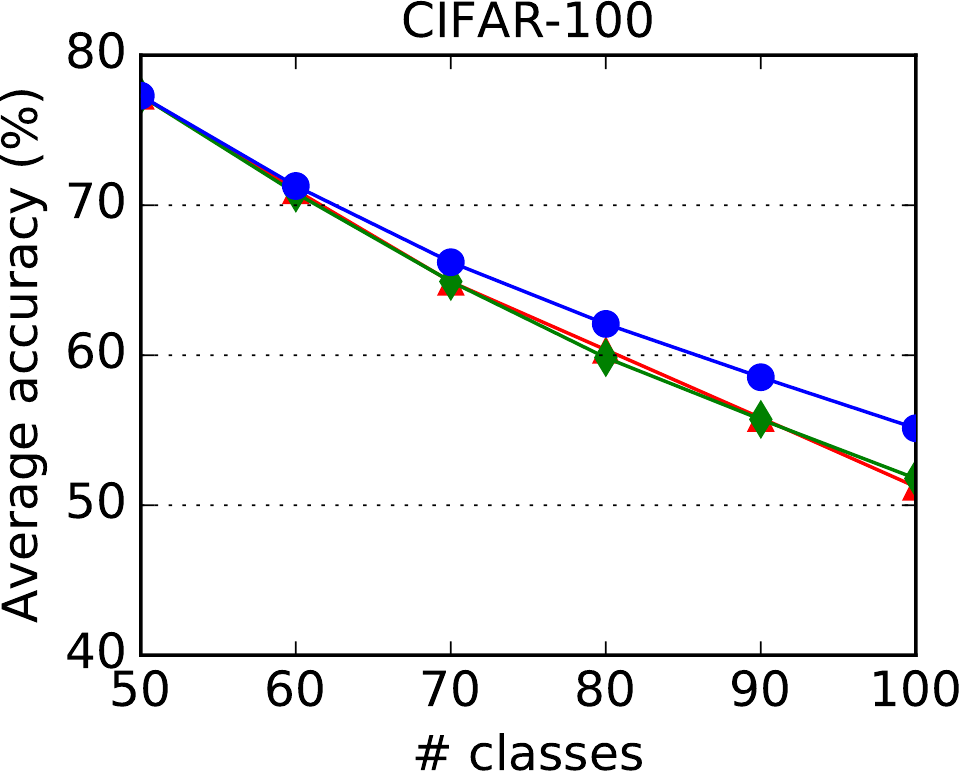}\\
(a) $N_D,N_R=500,0$ & (b) $N_D,N_R=500,20$\\
\end{tabular}
}\vspace{-.5em}
\caption{Comparison of three different methods of utilizing the synthetic samples from DF-GR in CIL on CIFAR-100. We used both DT-ID and DF-GR ($N_t,N_g=2,1$).\label{sec:exp:fig:05}}
\end{figure}

In Figure~\ref{sec:exp:fig:05}, we compare three different methods of utilizing the synthetic samples from DF-GR in CIL. The first method is our proposed DF-KD with weight imprinting to match the softmax output for both old and new classes. The second method is to match the softmax output for old classes only, which was used in \cite{li2017learning,wu2019large}. The third method is to use the target label of conditional generation directly to compute the cross-entropy $\expect_{p(z)p_i(y)}[\CE(y,f_i(g_i(z,y)))]$, instead of \eqref{sec:dtcil:dfgr:eq:02}. We observe that the proposed DF-KD with weight imprinting outperforms the other methods.

\setlength{\tabcolsep}{0.2em}
\begin{figure}[t]
\centering
{\small
\includegraphics[scale=.45]{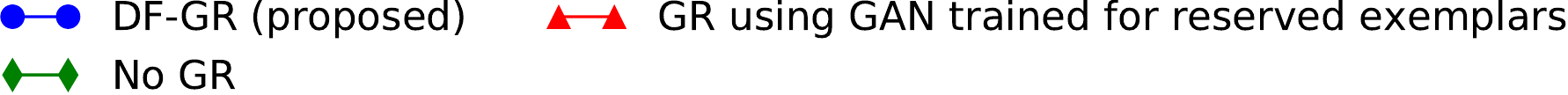}\\\vspace{.5em}
\begin{tabular}{cc}
\includegraphics[height=.37\columnwidth]{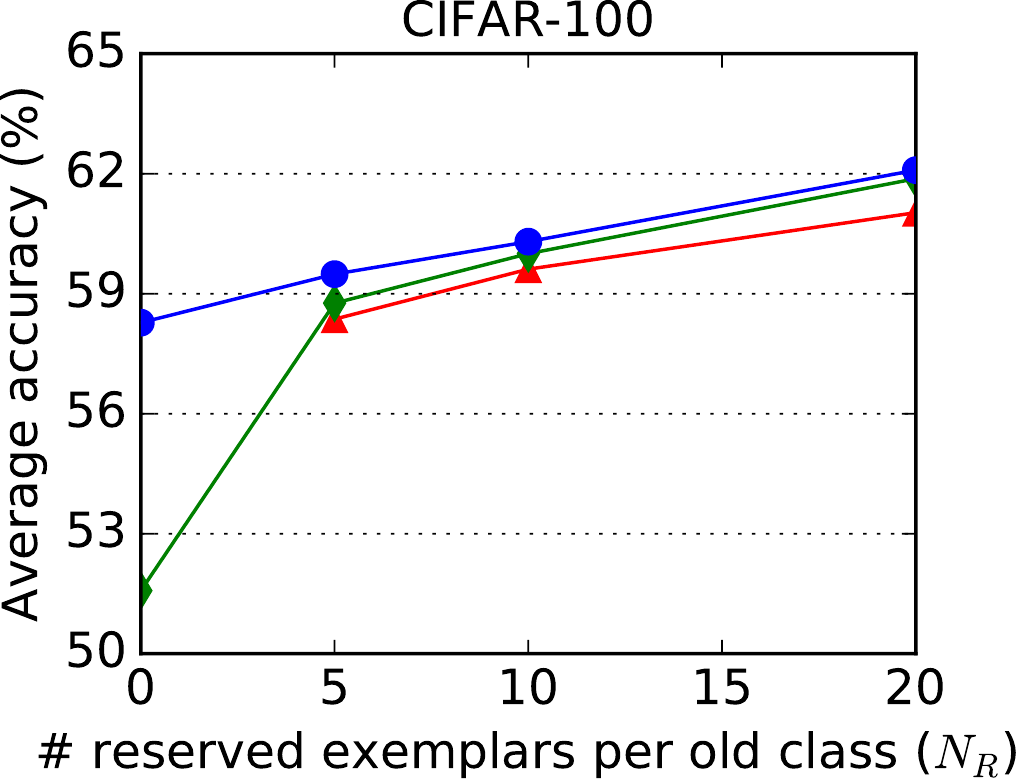}&
\includegraphics[height=.37\columnwidth]{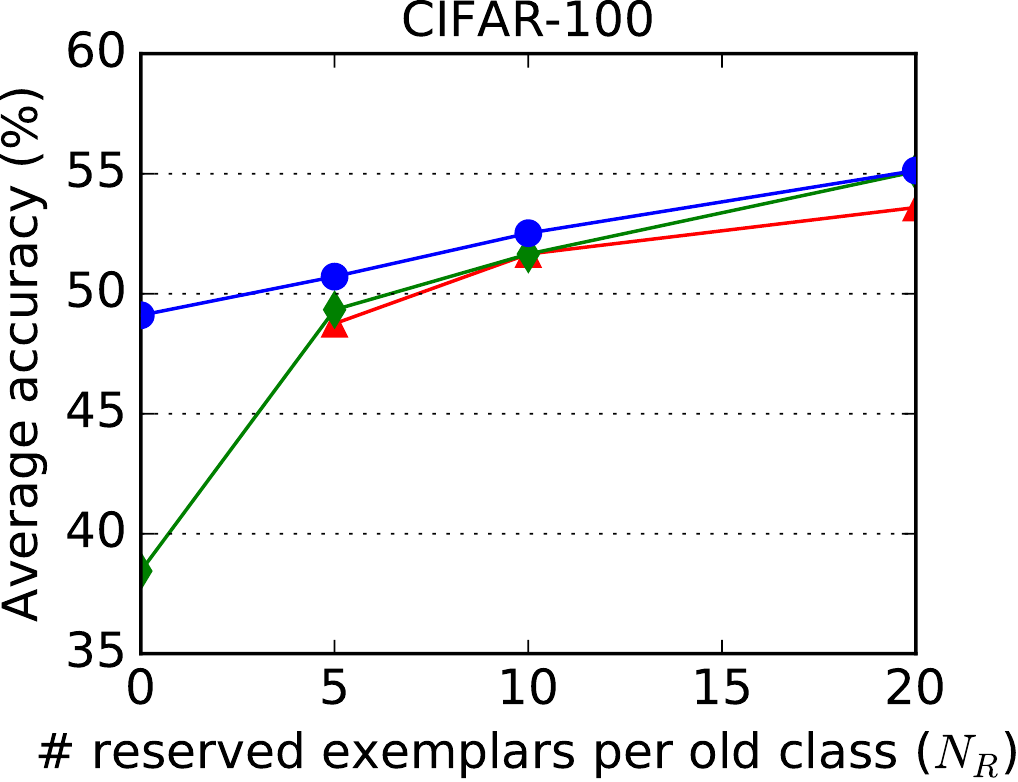}\\
(a) At time $3$ of CIL & (b) At time $5$ of CIL\\
\end{tabular}
}\vspace{-.5em}
\caption{Average accuracy for $80$ and $100$ seen classes at time~$3$ and $5$ of CIL on CIFAR-100, respectively, when using different numbers of reserved exemplars per old class. We used DT-ID in all cases ($N_t=2$).\label{sec:exp:fig:06}}
\end{figure}

In Figure~\ref{sec:exp:fig:06}, we investigate the impact of the number of reserved exemplars for old classes. In particular, we show the average accuracy for $80$ and $100$ seen classes at time~$3$ and $5$ of CIL on CIFAR-100, respectively. Recall that the major reason of proposing DF-GR is to curtail the burden of storing any past data (exemplars) or pre-trained generative models. We observe that the gain of using DF-GR is prominent when the number of reserved exemplars is small, and it shows the usefulness of the proposed DF-GR. We also compare our DF-GR to GR using the generative adversarial network (GAN) trained for reserved old exemplars only. The results show that our proposed DF-GR yields better CIL performance than GR using the GAN, when the training data is limited to a small number of reserved exemplars.

\setlength{\tabcolsep}{0.2em}
\begin{figure*}[t]
\centering
{\small
\includegraphics[scale=.45]{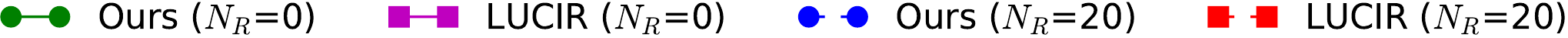}\\\vspace{.5em}
\begin{tabular}{ccc}
\includegraphics[height=.37\columnwidth]{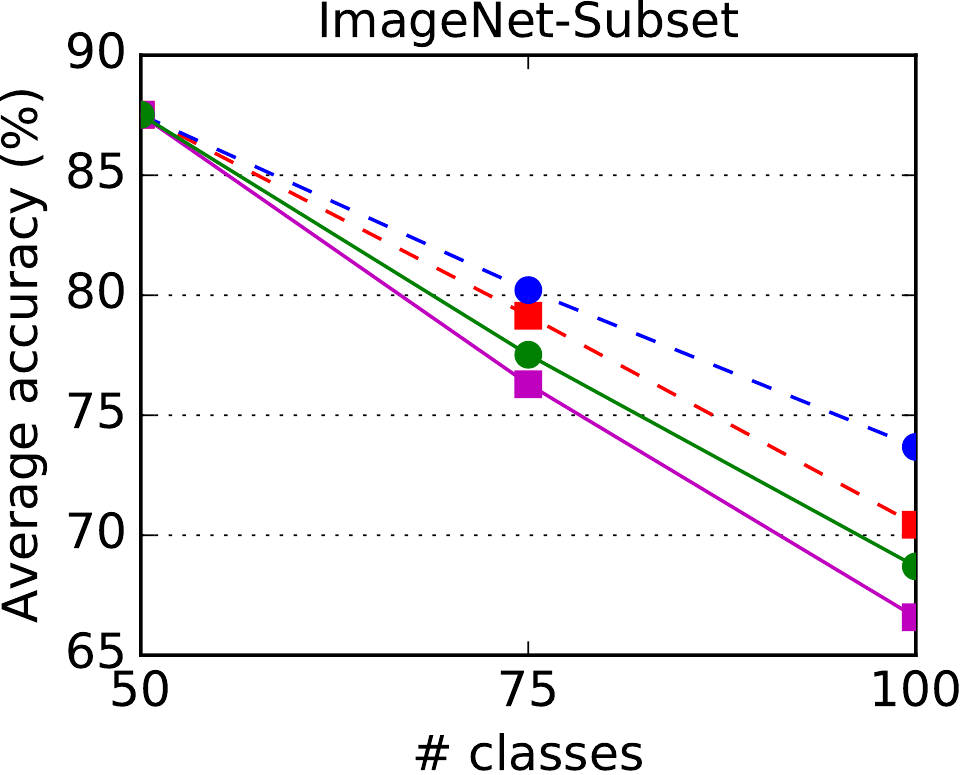}&
\includegraphics[height=.37\columnwidth]{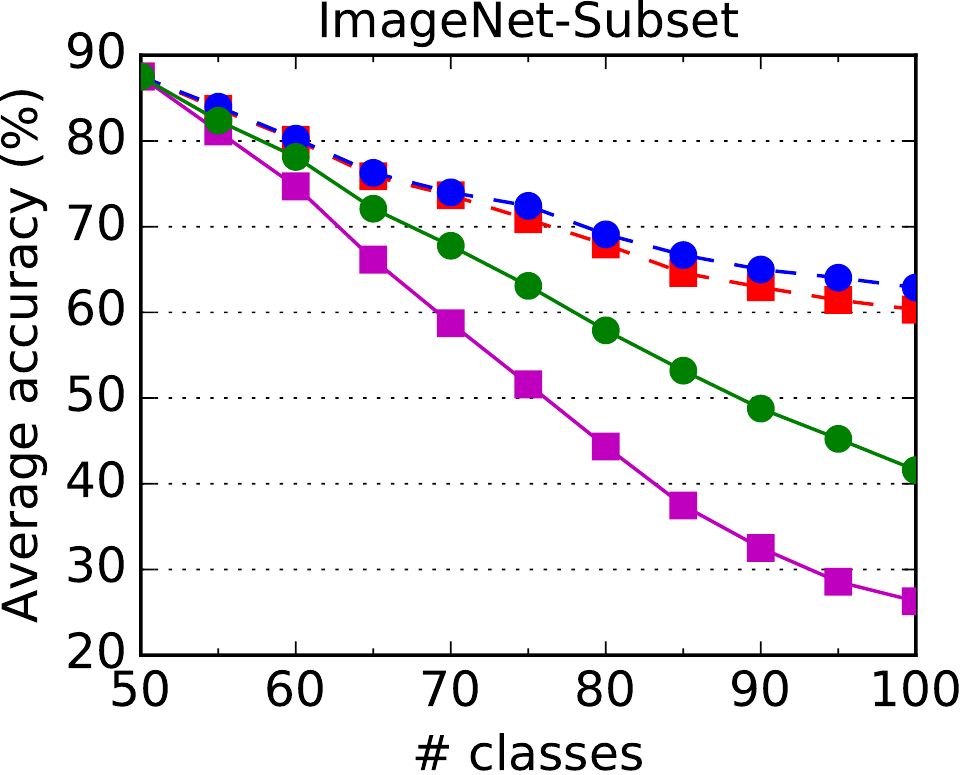}&
\includegraphics[height=.37\columnwidth]{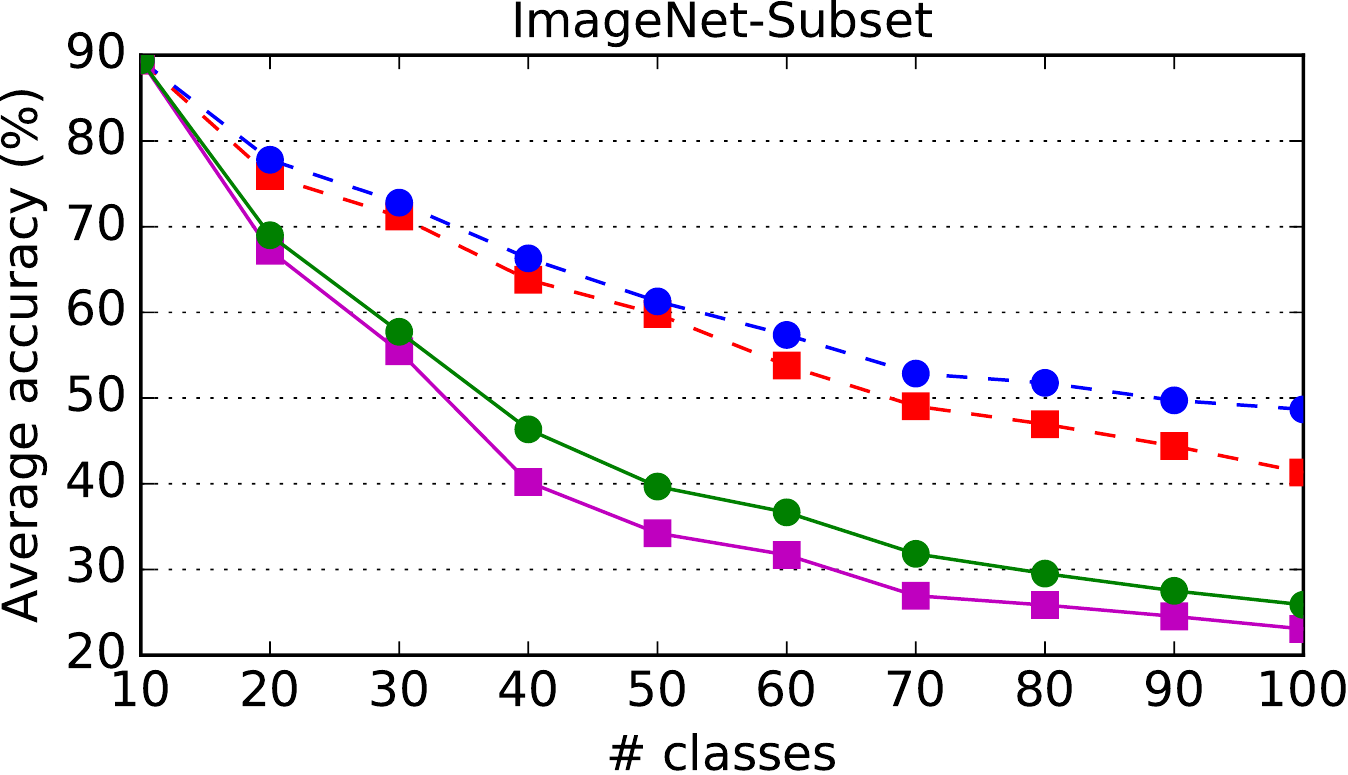}\\
(a) $50$ base classes + $25$ new classes per task  & (b) $50$ base classes + $5$ new classes per task  & (c) $10$ base classes + $10$ new classes per task\\
\end{tabular}
}\vspace{-.5em}
\caption{Average accuracy in CIL on ImageNet-Subset for different numbers of new classes added at each time of CIL with $50$ base classes (see (a) and (b)). We also show the average accuracy when CIL starts from $10$ base classes instead of $50$ base classes in (c). For ours, we used both DT-ID and DF-GR~($N_t,N_g=2,1$).\label{sec:exp:fig:04}}\vspace{-.5em}
\end{figure*}

In Figure~\ref{sec:exp:fig:04}(a,b), we show the average accuracy of CIL on ImageNet-Subset, when we change the number of new classes added at each time in CIL to $25$ and $5$, respectively. In Figure~\ref{sec:exp:fig:04}(c), we also show the average accuracy when CIL starts with $10$ base classes instead of $50$ base classes. Observe that our proposed method provides gains consistently over the baseline~\cite{hou2019learning} in all cases.

\subsection{Experimental results for data-limited CIL}\label{sec:exp:res:data-limited}

\setlength{\tabcolsep}{0.2em}
\begin{table}[t]
\centering
\caption{Average accuracy for all seen classes at each time of CIL on ImageNet-Subset in the data-limited CIL scenario, where we are not able to access the training data not only for old classes but also for new classes. We are given two pre-trained models for old and new classes with a small number of exemplars, respectively.\label{sec:exp:tbl:03}}\vspace{-.5em}
{\footnotesize
\begin{tabular}{ccccccccc}
\toprule
& $N_D,N_R$ & $N_t,N_g$ & \multicolumn{6}{c}{Average accuracy (\%) for seen classes}\\
\midrule
%\cmidrule{4-9}
%&           &           & Task0 & +Task1  & +Task2  & +Task3  & +Task4  & +Task5\\
Time
&           &           & 0 & 1 & 2 & 3 & 4 & 5\\
(\# classes)
&           &           & (50)  & (+10) & (+10) & (+10) & (+10) & (+10)\\
\midrule
\cite{hou2019learning}*
&20, 0      & 1, 0      & 87.52 & 79.40 & 70.91 & 61.35 & 52.09 & 43.00\\
\midrule
\multirow{2}{*}{Ours}
%&20, 0      & 2, 0      & 87.52 & 79.87 & 70.63 & 60.70 & 50.09 & 39.52\\
&20, 0      & 2, 1      & 87.52 & \textbf{80.17} & \textbf{74.54} & 68.50 & 62.53 & 56.80\\
&20, 0      & 2, 2      & 87.52 & 79.67 & 74.09 & \textbf{68.55} & \textbf{63.00} & \textbf{57.42}\\
\midrule
\cite{hou2019learning}*
&20, 20     & 1, 0      & 87.52 & 80.07 & 73.69 & 67.60 & 63.78 & 59.86\\
\midrule
\multirow{2}{*}{Ours}
%&20, 20     & 2, 0      & 87.52 & 80.23 & 74.14 & 69.53 & 65.42 & 62.18\\
&20, 20     & 2, 1      & 87.52 & 80.77 & \textbf{75.20} & 69.75 & 65.42 & 61.84\\
&20, 20     & 2, 2      & 87.52 & \textbf{81.17} & 75.14 & \textbf{69.88} & \textbf{65.73 }& \textbf{62.40}\\
%\midrule
%Ours
%&0, 0       & 2, 2      & 87.52 & 74.30 & 66.17 & 59.77 & 51.47 & 40.94\\
\bottomrule
\multicolumn{9}{r}{* Produced by us.}
\end{tabular}
}
\end{table}

\setlength{\tabcolsep}{0.2em}
\begin{figure}[t]
\centering
{\small
\includegraphics[scale=.45]{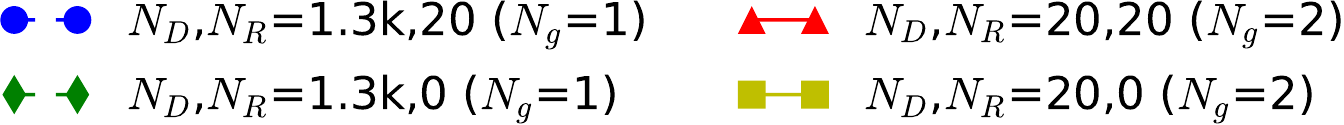}\\\vspace{.5em}
\begin{tabular}{cc}
\includegraphics[height=.37\columnwidth]{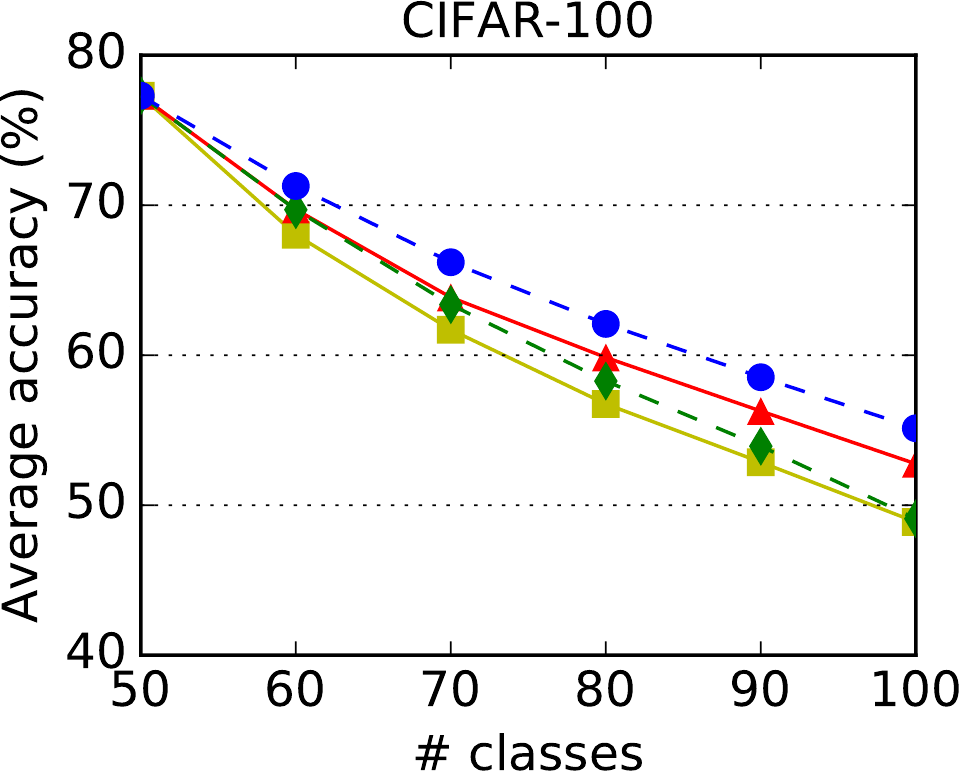}&
\includegraphics[height=.37\columnwidth]{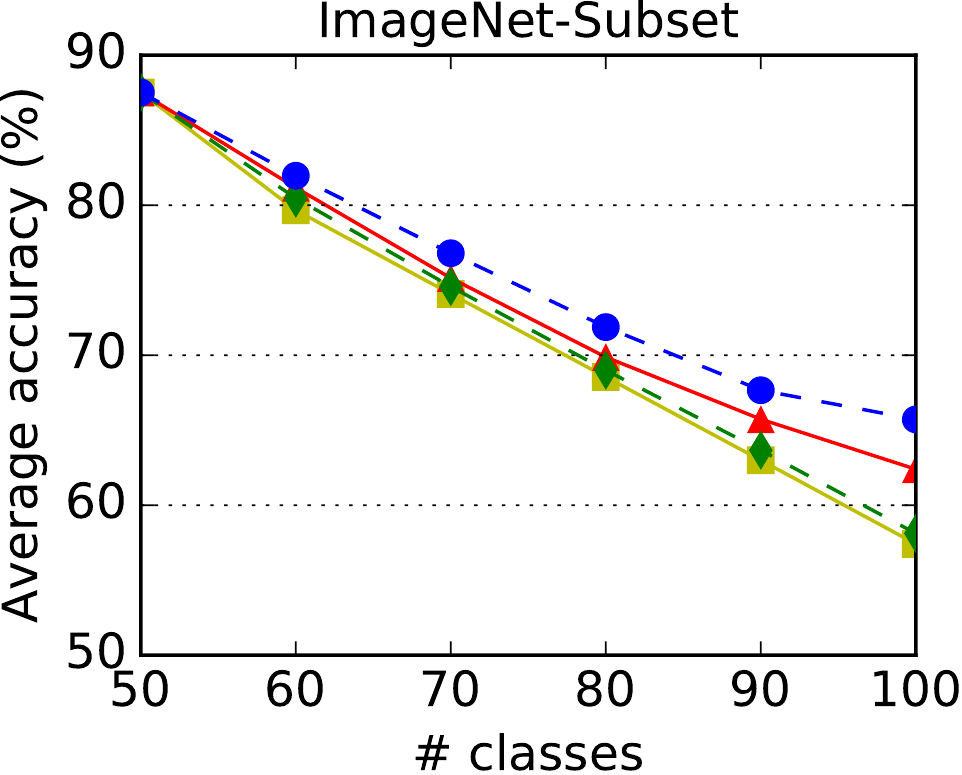}\\
(a) CIFAR-100 & (b) ImageNet-Subset\\
\end{tabular}
}\vspace{-.5em}
\caption{Average accuracy for different numbers of available data for old and new classes in DT-CIL on CIFAR-100 and ImageNet-Subset.\label{sec:exp:fig:03}}
\end{figure}

Data sharing becomes more and more difficult due to the large size and is often restricted to preserve data privacy. To adapt to this emerging challenge, we explore the data-limited CIL scenario, where data are limited not only for old classes but also for new classes. In the data-limited CIL scenario, instead of sharing a large number of data, we assume that two pre-trained models are provided for CIL. The first one is the model pre-trained for old classes, as in the conventional CIL. The second one is the pre-trained model on the data for new classes. We employ our proposed DF-GR to reproduce not only the data for old classes but also the data for new classes. That is, we train two generators based on two pre-trained models provided for data-limited CIL, respectively. Then, we add the synthetics samples from two generators when computing the DT-ID and LF losses in \eqref{sec:dtcil:dtid:eq:03} and \eqref{sec:cil:eq:03}, similar to the single DF-GR case (see Section~\ref{sec:dtcil:dfgr}). In additional to that, we have two DF-KD losses from dual DF-GR for old and new classes, respectively. For the synthetic samples from the generator for old classes, the DF-KD loss is computed after weight imprinting, as described in Section~\ref{sec:dtcil:dfgr}. For the synthetic samples from the generator for new classes, we only match the softmax output for new classes, without weight imprinting, in DF-KD, since weight imprinting for a large number of old classes was not effective in our trials.

For the data-limited CIL scenario, we perform experiments on two cases of $N_D,N_R=20,20$ and $N_D,N_R=20,0$ (see Table~\ref{sec:exp:tbl:01}). Table~\ref{sec:exp:tbl:03} shows the results for these two cases on ImageNet-Subset. Observe that our proposed DT-CIL with DT-ID and DF-GR outperforms the baseline LUCIR~\cite{hou2019learning}. In Figure~\ref{sec:exp:fig:03}, we compare the average accuracy for different numbers of available data for old and new classes. Observe that the average accuracy of the data-limited CIL with a small number of exemplars approaches to the conventional CIL performance.

\section{Conclusion}\label{sec:conclusion}

In this paper, we proposed a novel dual-teacher knowledge transfer method with data-free generative replay for class-incremental learning. First, DF-GR was used to produce the synthetic samples for old and even for new classes. No original training data and no pre-trained generative models are used in DF-GR. Second, we introduced DT-ID to transfer knowledge from two pre-trained models for old and new classes to one fused model. Our proposed dual-teacher CIL deploys both DF-GR and DT-ID. In our experiments, we implemented the proposed schemes on top of one of the state-of-the-art CIL methods and showed their gains. Last but not least, we demonstrated the potential of DF-GR for data-limited CIL, where available training data are limited due to the cost of data sharing and to preserve data privacy.

{\small
\bibliographystyle{ieee_fullname}
\bibliography{ref}

\begin{thebibliography}{10}\itemsep=-1pt

\bibitem{ahn2019variational}
Sungsoo Ahn, Shell~Xu Hu, Andreas Damianou, Neil~D Lawrence, and Zhenwen Dai.
\newblock Variational information distillation for knowledge transfer.
\newblock In {\em Proceedings of the IEEE Conference on Computer Vision and
  Pattern Recognition}, pages 9163--9171, 2019.

\bibitem{aljundi2018memory}
Rahaf Aljundi, Francesca Babiloni, Mohamed Elhoseiny, Marcus Rohrbach, and
  Tinne Tuytelaars.
\newblock Memory aware synapses: Learning what (not) to forget.
\newblock In {\em Proceedings of the European Conference on Computer Vision},
  pages 139--154, 2018.

\bibitem{barber2003algorithm}
David Barber and Felix~V Agakov.
\newblock The {IM} algorithm: A variational approach to information
  maximization.
\newblock In {\em Advances in Neural Information Processing Systems}, page
  None, 2003.

\bibitem{castro2018end}
Francisco~M Castro, Manuel~J Mar{\'\i}n-Jim{\'e}nez, Nicol{\'a}s Guil, Cordelia
  Schmid, and Karteek Alahari.
\newblock End-to-end incremental learning.
\newblock In {\em Proceedings of the European Conference on Computer Vision},
  pages 233--248, 2018.

\bibitem{chaudhry2018riemannian}
Arslan Chaudhry, Puneet~K Dokania, Thalaiyasingam Ajanthan, and Philip~HS Torr.
\newblock Riemannian walk for incremental learning: Understanding forgetting
  and intransigence.
\newblock In {\em Proceedings of the European Conference on Computer Vision},
  pages 532--547, 2018.

\bibitem{choi2020data}
Yoojin Choi, Jihwan Choi, Mostafa El-Khamy, and Jungwon Lee.
\newblock Data-free network quantization with adversarial knowledge
  distillation.
\newblock In {\em Proceedings of the IEEE Conference on Computer Vision and
  Pattern Recognition Workshops}, pages 710--711, 2020.

\bibitem{cover2012elements}
Thomas~M Cover and Joy~A Thomas.
\newblock {\em Elements of Information Theory}.
\newblock John Wiley \& Sons, 2012.

\bibitem{de2017modulating}
Harm De~Vries, Florian Strub, J{\'e}r{\'e}mie Mary, Hugo Larochelle, Olivier
  Pietquin, and Aaron~C Courville.
\newblock Modulating early visual processing by language.
\newblock In {\em Advances in Neural Information Processing Systems}, pages
  6594--6604, 2017.

\bibitem{dhar2019learning}
Prithviraj Dhar, Rajat~Vikram Singh, Kuan-Chuan Peng, Ziyan Wu, and Rama
  Chellappa.
\newblock Learning without memorizing.
\newblock In {\em Proceedings of the IEEE Conference on Computer Vision and
  Pattern Recognition}, pages 5138--5146, 2019.

\bibitem{goodfellow2013empirical}
Ian~J Goodfellow, Mehdi Mirza, Da Xiao, Aaron Courville, and Yoshua Bengio.
\newblock An empirical investigation of catastrophic forgetting in
  gradient-based neural networks.
\newblock {\em arXiv preprint arXiv:1312.6211}, 2013.

\bibitem{he2016deep}
Kaiming He, Xiangyu Zhang, Shaoqing Ren, and Jian Sun.
\newblock Deep residual learning for image recognition.
\newblock In {\em Proceedings of the IEEE Conference on Computer Vision and
  Pattern Recognition}, pages 770--778, 2016.

\bibitem{hinton2015distilling}
Geoffrey Hinton, Oriol Vinyals, and Jeff Dean.
\newblock Distilling the knowledge in a neural network.
\newblock {\em arXiv preprint arXiv:1503.02531}, 2015.

\bibitem{hou2019learning}
Saihui Hou, Xinyu Pan, Chen~Change Loy, Zilei Wang, and Dahua Lin.
\newblock Learning a unified classifier incrementally via rebalancing.
\newblock In {\em Proceedings of the IEEE Conference on Computer Vision and
  Pattern Recognition}, pages 831--839, 2019.

\bibitem{kemker2018fearnet}
Ronald Kemker and Christopher Kanan.
\newblock {FearNet}: Brain-inspired model for incremental learning.
\newblock In {\em International Conference on Learning Representations}, 2018.

\bibitem{kingma2014adam}
Diederik Kingma and Jimmy Ba.
\newblock Adam: A method for stochastic optimization.
\newblock In {\em International Conference on Learning Representations}, 2015.

\bibitem{kirkpatrick2017overcoming}
James Kirkpatrick, Razvan Pascanu, Neil Rabinowitz, Joel Veness, Guillaume
  Desjardins, Andrei~A Rusu, Kieran Milan, John Quan, Tiago Ramalho, Agnieszka
  Grabska-Barwinska, et~al.
\newblock Overcoming catastrophic forgetting in neural networks.
\newblock {\em Proceedings of the National Academy of Sciences},
  114(13):3521--3526, 2017.

\bibitem{krizhevsky2009learning}
Alex Krizhevsky.
\newblock Learning multiple layers of features from tiny images.
\newblock {\em Technical report, Univ. of Toronto}, 2009.

\bibitem{li2017learning}
Zhizhong Li and Derek Hoiem.
\newblock Learning without forgetting.
\newblock {\em IEEE Transactions on Pattern Analysis and Machine Intelligence},
  40(12):2935--2947, 2017.

\bibitem{liu2020generative}
Xialei Liu, Chenshen Wu, Mikel Menta, Luis Herranz, Bogdan Raducanu, Andrew~D
  Bagdanov, Shangling Jui, and Joost van~de Weijer.
\newblock Generative feature replay for class-incremental learning.
\newblock In {\em Proceedings of the IEEE Conference on Computer Vision and
  Pattern Recognition Workshops}, pages 226--227, 2020.

\bibitem{liu2020mnemonics}
Yaoyao Liu, Yuting Su, An-An Liu, Bernt Schiele, and Qianru Sun.
\newblock Mnemonics training: Multi-class incremental learning without
  forgetting.
\newblock In {\em Proceedings of the IEEE Conference on Computer Vision and
  Pattern Recognition}, pages 12245--12254, 2020.

\bibitem{miyato2018cgans}
Takeru Miyato and Masanori Koyama.
\newblock {cGANs} with projection discriminator.
\newblock In {\em International Conference on Learning Representations}, 2018.

\bibitem{nesterov1983method}
Yurii Nesterov.
\newblock A method for unconstrained convex minimization problem with the rate
  of convergence {$O(1/k^2)$}.
\newblock In {\em Doklady AN USSR}, volume 269, pages 543--547, 1983.

\bibitem{nguyen2018variational}
Cuong~V Nguyen, Yingzhen Li, Thang~D Bui, and Richard~E Turner.
\newblock Variational continual learning.
\newblock In {\em International Conference on Learning Representations}, 2018.

\bibitem{parisi2019continual}
German~I Parisi, Ronald Kemker, Jose~L Part, Christopher Kanan, and Stefan
  Wermter.
\newblock Continual lifelong learning with neural networks: A review.
\newblock {\em Neural Networks}, 113:54--71, 2019.

\bibitem{qi2018low}
Hang Qi, Matthew Brown, and David~G Lowe.
\newblock Low-shot learning with imprinted weights.
\newblock In {\em Proceedings of the IEEE Conference on Computer Vision and
  Pattern Recognition}, pages 5822--5830, 2018.

\bibitem{rebuffi2017icarl}
Sylvestre-Alvise Rebuffi, Alexander Kolesnikov, Georg Sperl, and Christoph~H
  Lampert.
\newblock {iCaRL}: Incremental classifier and representation learning.
\newblock In {\em Proceedings of the IEEE Conference on Computer Vision and
  Pattern Recognition}, pages 2001--2010, 2017.

\bibitem{ronneberger2015u}
Olaf Ronneberger, Philipp Fischer, and Thomas Brox.
\newblock {U-Net}: Convolutional networks for biomedical image segmentation.
\newblock In {\em International Conference on Medical Image Computing and
  Computer-Assisted Intervention}, pages 234--241. Springer, 2015.

\bibitem{russakovsky2015imagenet}
Olga Russakovsky, Jia Deng, Hao Su, Jonathan Krause, Sanjeev Satheesh, Sean Ma,
  Zhiheng Huang, Andrej Karpathy, Aditya Khosla, Michael Bernstein, et~al.
\newblock {ImageNet} large scale visual recognition challenge.
\newblock {\em International Journal of Computer Vision}, 115(3):211--252,
  2015.

\bibitem{shin2017continual}
Hanul Shin, Jung~Kwon Lee, Jaehong Kim, and Jiwon Kim.
\newblock Continual learning with deep generative replay.
\newblock In {\em Advances in Neural Information Processing Systems}, pages
  2990--2999, 2017.

\bibitem{ulyanov2018deep}
Dmitry Ulyanov, Andrea Vedaldi, and Victor Lempitsky.
\newblock Deep image prior.
\newblock In {\em Proceedings of the IEEE Conference on Computer Vision and
  Pattern Recognition}, pages 9446--9454, 2018.

\bibitem{wu2019large}
Yue Wu, Yinpeng Chen, Lijuan Wang, Yuancheng Ye, Zicheng Liu, Yandong Guo, and
  Yun Fu.
\newblock Large scale incremental learning.
\newblock In {\em Proceedings of the IEEE Conference on Computer Vision and
  Pattern Recognition}, pages 374--382, 2019.

\bibitem{xiang2019incremental}
Ye Xiang, Ying Fu, Pan Ji, and Hua Huang.
\newblock Incremental learning using conditional adversarial networks.
\newblock In {\em Proceedings of the IEEE International Conference on Computer
  Vision}, pages 6619--6628, 2019.

\bibitem{zenke2017continual}
Friedemann Zenke, Ben Poole, and Surya Ganguli.
\newblock Continual learning through synaptic intelligence.
\newblock In {\em International Conference on Machine Learning}, pages
  3987--3995, 2017.

\end{thebibliography}
}
\onecolumn

\ifmain
\appendixpage
\else
%%%%%%%%% TITLE
\title{Dual-Teacher Class-Incremental Learning With Data-Free Generative Replay\\--- \emph{Supplementary Materials} ---}

%\author{Yoojin Choi, Mostafa El-Khamy, Jungwon Lee\\
%SoC R\&D, Samsung Semiconductor Inc., San Diego, CA 92121, USA\\
%{\tt\small \{yoojin.c,mostafa.e,jungwon2.lee\}@samsung.com}
%}
\author{%Yoojin Choi\Mark{1}, Mostafa El-Khamy\Mark{1}, Jungwon Lee\Mark{2}\\
\begin{tabular}{ccc}
Yoojin Choi, Mostafa El-Khamy & \qquad\qquad & Jungwon Lee\\
SoC R\&D, Samsung Semiconductor Inc. & & System LSI, Samsung Electronics\\
San Diego, CA 92121, USA & & South Korea\\
{\tt\small \{yoojin.c,mostafa.e\}@samsung.com} & & {\tt\small jungwon2.lee@samsung.com}\\
\end{tabular}
}

\maketitle
\thispagestyle{empty}
\fi

\setcounter{section}{0}
\renewcommand*{\thesection}{\Alph{section}}
\renewcommand*{\theHsection}{app.\the\value{section}}

\section{Experiments}\label{app:exp}

\subsection{Training procedure}\label{app:exp:training}

At time~$0$ of CIL on CIFAR-100, the ResNet-32 model is trained with Nesterov's accelerated gradient (NAG)~\cite{nesterov1983method} of batch size~$128$ for $160$ epochs by using the original CIFAR-100 data of randomly selected $50$ base classes. The learning rate starts from $0.1$ and is reduced to $0.01$ and $0.001$ at epoch $80$ and $120$, respectively. At time $i\geq1$, to re-train the ResNet-32 model for CIL on additional CIFAR-100 classes, we use NAG of batch size~$256$ for $160$ epochs, where each epoch consists of $50$ batches. The learning rate starts from $0.1$ and is reduced to $0.01$ and $0.001$ at epoch $80$ and $120$, respectively. 

At time~$0$ of CIL on ImageNet-Subset and ImageNet-Full, the ResNet-18 models are trained with NAG of batch size~$128$ for $90$ epochs by using the original ImageNet data of randomly selected $50$ and $500$ base classes, respectively. The learning rate starts from $0.1$ and is reduced to $0.01$ and $0.001$ at epoch $30$ and $60$, respectively. At time~$i\geq1$, for CIL on ImageNet, we re-train the ResNet-18 models with NAG of batch size~$100$ for $90$ epochs, where each epoch consists of $128$ and $1280$ batches for ImageNet-Subset and ImageNet-Full, respectively. The learning rate starts from $0.1$ and is reduced to $0.01$ and $0.001$ at epoch $30$ and $60$, respectively.

At time~$i\geq1$, for CIL, the original (or exemplary) and synthetic samples for old and new classes are mixed in each batch by the following distribution, depending on their availability, where $B$ denotes the batch size (see Table~\ref{sec:exp:tbl:01} for the notations of $N_D$, $N_R$, and $N_g$).
\setlength{\tabcolsep}{0.5em}
\begin{table}[h]
\centering
%\caption{Distribution of the original (or exemplary) and synthetic samples for old and new classes in each batch of size~$B$, depending on their availability.\label{sec:suppl:tbl:01}}\vspace{-.5em}
{\footnotesize
\begin{tabular}{rrc|cccc}
\toprule
$N_D$ & $N_R$ & $N_g$ & \multicolumn{2}{c}{New classes} & \multicolumn{2}{c}{Old classes}\\
 & & & Original (or exemplary) & Synthetic & Exemplary & Synthetic\\
\midrule
\multirow{3}{*}{$>0$} & \multirow{3}{*}{$>0$}
 & $0$ & $B/2$ & $0$ & $B/2$ & $0$\\
 & & $1$ & $B/2$ & $0$ & $B/4$ & $B/4$\\
 & & $2$ & $B/4$ & $B/4$ & $B/4$ & $B/4$\\
\midrule
\multirow{3}{*}{$>0$} & \multirow{3}{*}{$0$}
 & $0$ & $B$ & $0$ & $0$ & $0$\\
 & & $1$ & $B/2$ & $0$ & $0$ & $B/2$\\
 & & $2$ & $B/4$ & $B/4$ & $0$ & $B/2$\\
%\midrule
%$0$ & $0$ & $2$ & $0$ & $B/2$ & $0$ & $B/2$\\
\bottomrule
\end{tabular}
}
\end{table}

For the factor~$\alpha_i$ in front of the less-forget (LF) loss in \eqref{sec:dtcil:dtid:eq:04}, we set
\[
\alpha_i=\alpha_0\sqrt{\frac{|C_0^{i-1}|}{|C_i|}},
\]
with $\alpha_0=5$ for CIFAR-100 and $\alpha_0=10$ for ImageNet in the conventional CIL scenario, as suggested in \cite{hou2019learning}, where $|C|$ denotes the size of set $C$. If we have no exemplars for old classes, we double $\alpha_0$ to enforce a stronger less-forget constraint. In the data-limited CIL scenario, we search the best $\alpha_0$ in $\{1,5,10\}$ for CIFAR-100 and in $\{2,10,20\}$ for ImageNet. We also use gradient clipping to avoid exploding gradients, as suggested in \cite{ahn2019variational}---we clip the norm of gradients by $1$.

For ImageNet-Full, we additionally use the margin ranking loss for inter-class separation and also perform class-balanced fine-tuning at the end of each incremental training, if there are reserved exemplars for old classes, in ours as well as when we reproduce the baseline LUCIR~\cite{hou2019learning}. Note that those two components were shown to be effective significantly for ImageNet-Full in \cite{hou2019learning}. For CIFAR-100 and ImageNet-Subset, we do not use those two additional components either in ours or when we reproduce the baseline LUCIR~\cite{hou2019learning}.

%\subsection{Evaluation metrics}\label{app:exp:metric}
%
%\textbf{Average accuracy}. Let $A_i(c)$ be the average accuracy for the test data of class~$c\in C_0^i$ at time~$i$. The average accuracy at time~$i$ is given by
%\[
%A_i=\frac{1}{|C_0^i|}\sum_{c\in C_0^i}A_i(c).
%\]
%
%\textbf{Average forgetting}. The average forgetting is defined as the difference between the maximum knowledge gained about a class throughout the learning process in the past and the knowledge the model currently has about it. This, in turn, gives an estimate of how much the model forgot about the class given its current state~\cite{chaudhry2018riemannian}. At time~$i$, we define 
%\[
%F_i(c)=\max_{0\leq j\leq i-1}\{A_j(c)-A_i(c)\},
%\ \ \
%c\in C_0^{i-1},
%\]
%and the average forgetting is given by their average over all old classes in $C_0^{i-1}$ as below:
%\[
%F_i=\frac{1}{|C_0^{i-1}|}\sum_{c\in C_0^{i-1}}F_i(c).
%\]

\subsection{Generator architecture}\label{app:exp:gen}

\begin{figure*}[t]
\centering
\includegraphics[height=.2\textwidth]{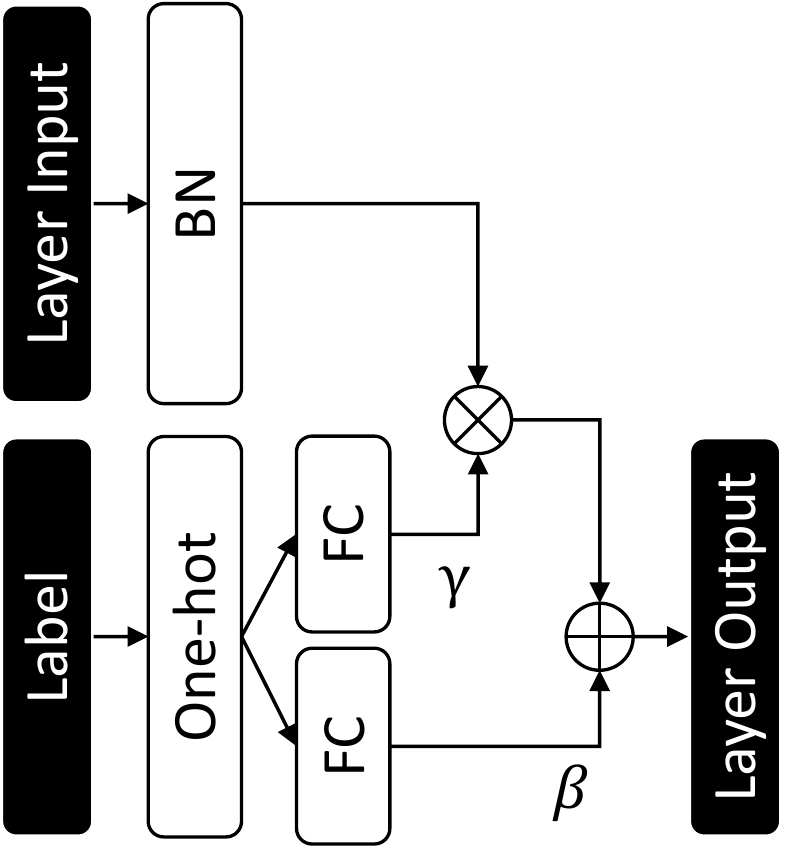}
\caption{Conditional batch normalization. \textbf{BN} denotes a vanilla batch normalization layer without scaling and shifting after normalization. \textbf{One-hot} implies one-hot encoding. \textbf{FC} denotes a fully-connected layer. The symbols~$\bigotimes$ and $\bigoplus$ stand for channel-wise scaling and channel-wise bias addition, respectively.\label{app:gen:fig:01}}%\vspace{-1em}
\end{figure*}

\begin{figure*}[t]
\centering
\includegraphics[height=.15\textwidth]{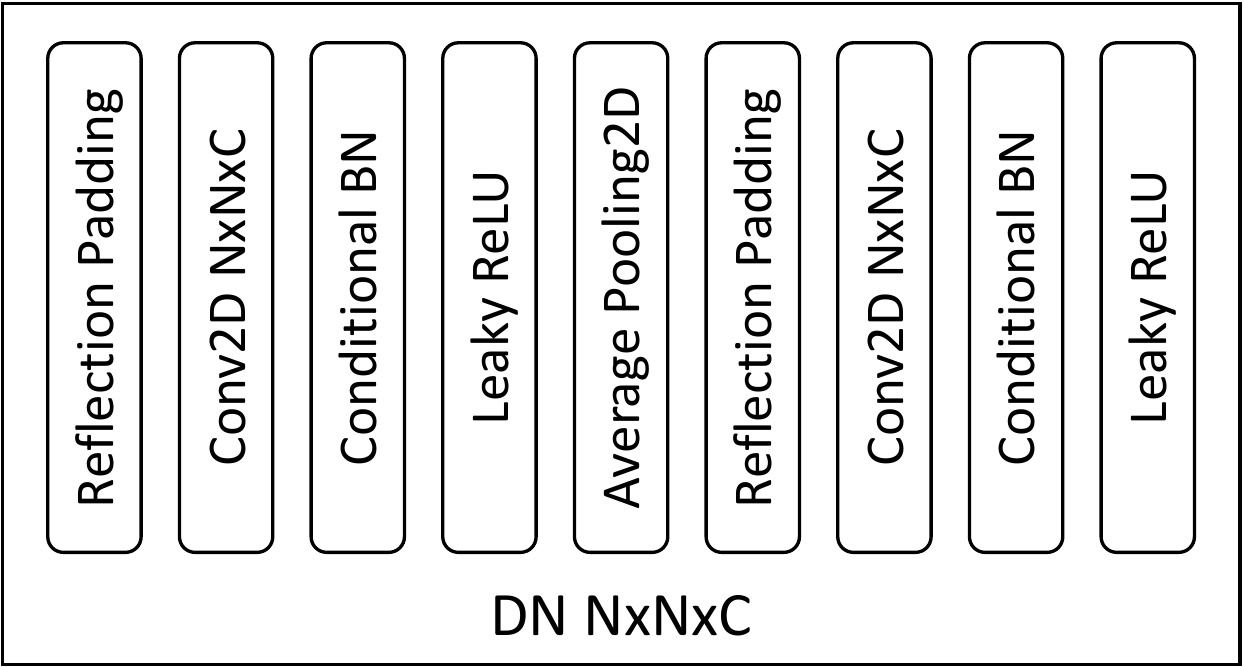}\quad
\includegraphics[height=.15\textwidth]{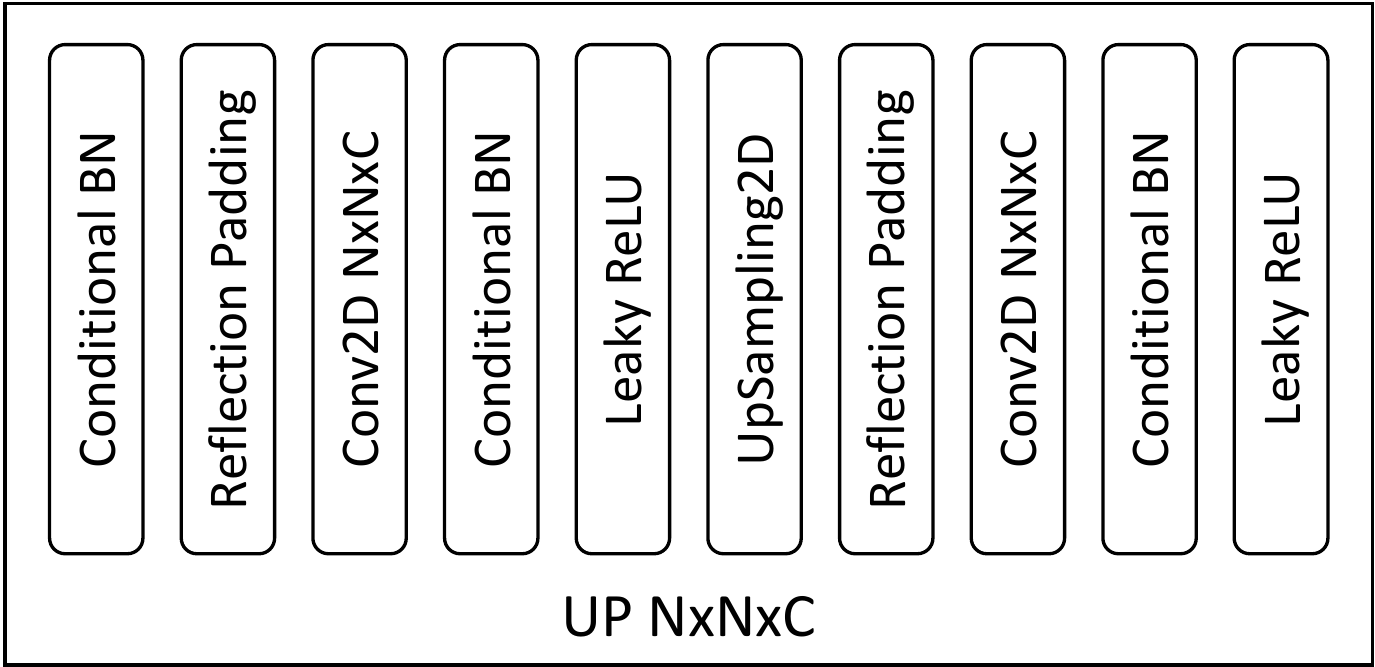}\quad
\includegraphics[height=.15\textwidth]{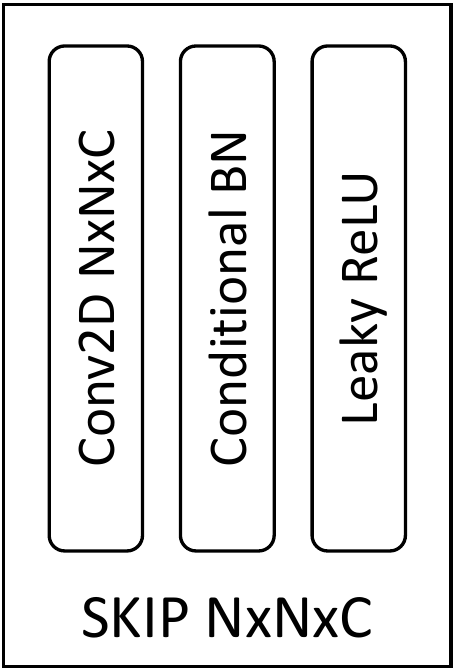}\\\vspace{2em}
\includegraphics[scale=.38]{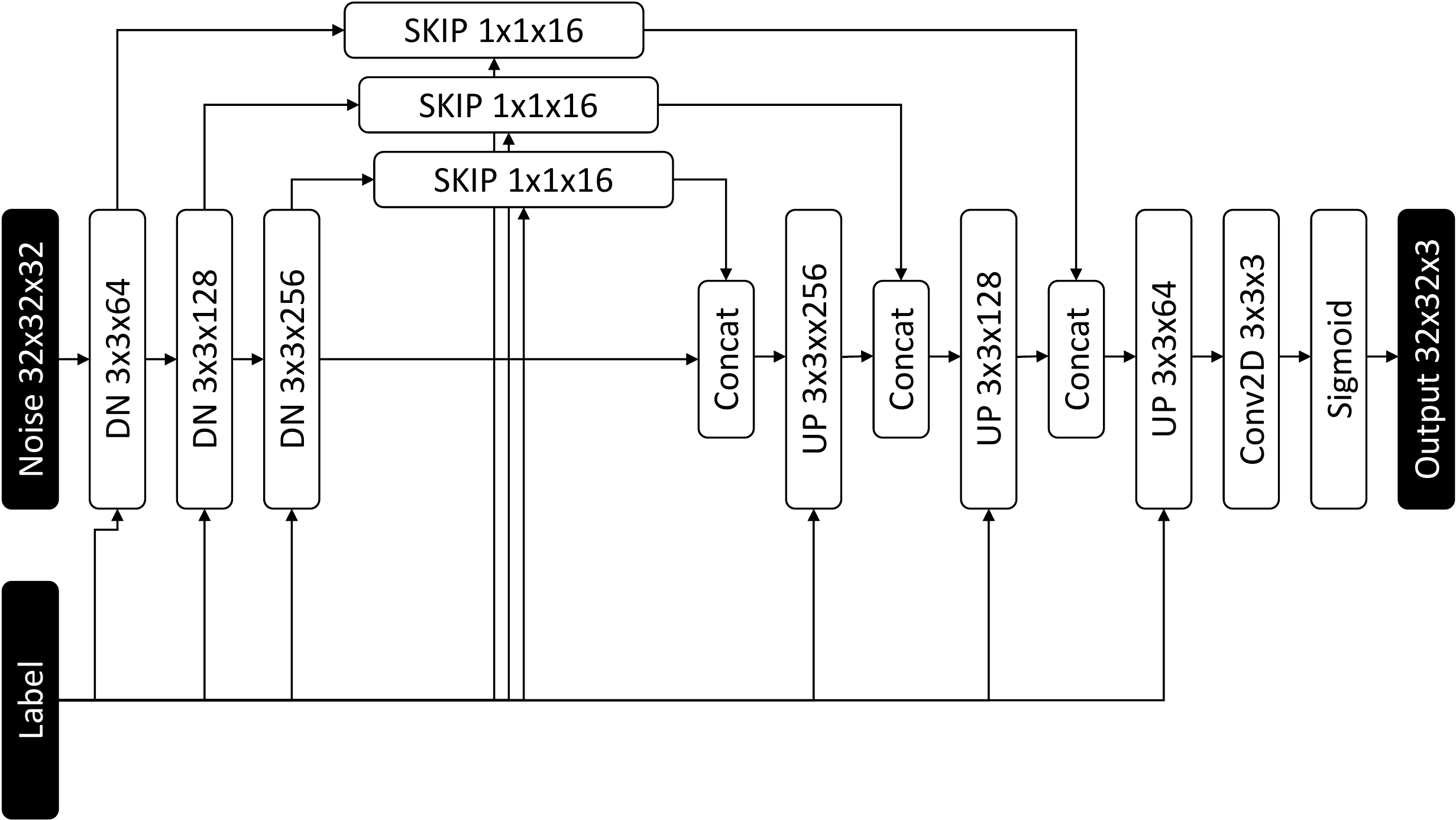}\\
(a) Generator for DF-GR in CIL on CIFAR-100\\\vspace{2em}
\includegraphics[scale=.38]{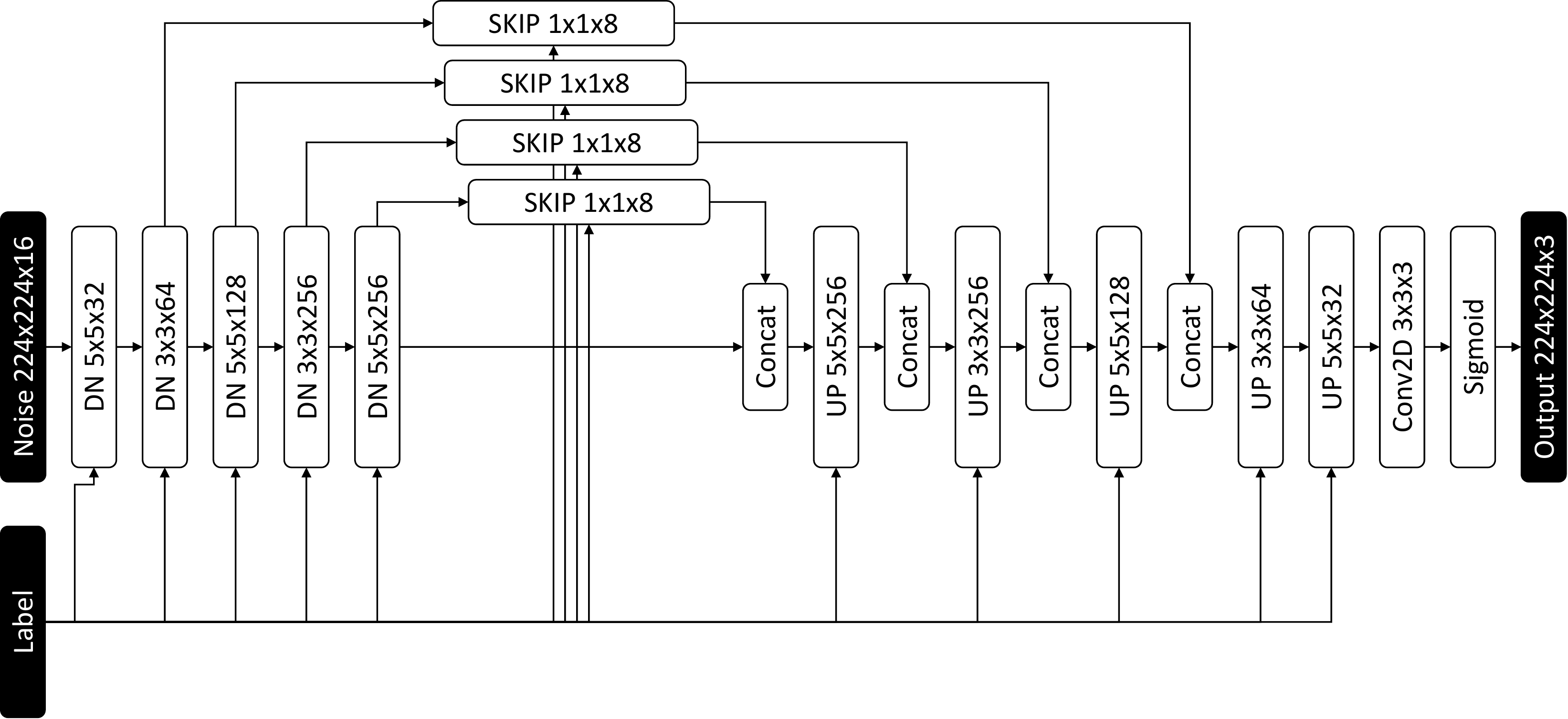}\\
(b) Generator for DF-GR in CIL on ImageNet-Subset and ImageNet-Full\\
\caption{Generator architectures for DF-GR in CIL on CIFAR-100 and ImageNet. \textbf{Conv2D NxNxC} denotes a convolutional layer of kernel size~$N\times N$ with $C$ output channels. \textbf{Reflection Padding} implies that we add reflection padding to make the input size (width and height) and the output size of the following convolution be the same. For \textbf{UpSampling2D}, we employ bilinear upsampling. \textbf{Conditional BN} can be found in Figure~\ref{app:gen:fig:01}. \textbf{Concat} stands for concatenation. \textbf{Noise NxNxC} denotes the input random noise of shape~$N\times N\times C$, sampled from the standard normal distribution.\label{app:gen:fig:03}}%\vspace{-1em}
\end{figure*}

In Figure~\ref{app:gen:fig:01}, we illustrate conditional batch normalization (BN) used in our conditional generators for DF-GR. Observe that the channel-wise scaling factor~$\gamma$ and the channel-wise bias~$\beta$ depend on the input label (condition). In Figure~\ref{app:gen:fig:03}, we depict the generator architectures used in our experiments for DF-GR in CIL on CIFAR-100 and ImageNet, respectively.

\end{document}